%% file: aaai25.tex
\documentclass[letterpaper]{article} 
\usepackage{aaai25}  
\usepackage{times}  
\usepackage{helvet}  
\usepackage{courier}  
\usepackage[hyphens]{url}  
\usepackage{graphicx} 
\usepackage{natbib}  
\usepackage{caption} 
\usepackage{algorithm}
\usepackage{algorithmic}
\usepackage{xcolor}
\usepackage{amsmath}
\usepackage{multirow}
\usepackage{subfigure}
\usepackage{amssymb}
\usepackage[switch]{lineno}
\newcommand{\subsubsubsection}[1]{\paragraph{#1}\mbox{}\\}
\usepackage{newfloat}
\usepackage{listings}
\DeclareCaptionStyle{ruled}{labelfont=normalfont,labelsep=colon,strut=off} 
\floatstyle{ruled}
\newfloat{listing}{tb}{lst}{}
\floatname{listing}{Listing}
\title{StoryWeaver: A Unified World Model for \\Knowledge-Enhanced Story Character Customization}
\author {
    Jinlu Zhang \equalcontrib \textsuperscript{\rm 1},
    Jiji Tang \equalcontrib \textsuperscript{\rm 2},
    Rongsheng Zhang \textsuperscript{\rm 2},
    Tangjie Lv \textsuperscript{\rm 2},
    Xiaoshuai Sun \thanks{*Corresponding author} \textsuperscript{\rm 1}
}
\affiliations {
    \textsuperscript{\rm 1}Key Laboratory of Multimedia Trusted Perception and Efficient Computing,\\ Ministry of Education of China, Xiamen University, 361005, P.R. China.\\
    \textsuperscript{\rm 2}Fuxi AI Lab, Netease Inc.\\
    zhangjinlu@stu.xmu.edu.cn, \{tangjiji01,zhangrongsheng,hzlvtangjie\}@corp.netease.com, xssun@xmu.edu.cn
}

\usepackage{bibentry}
\let\oldtwocolumn\twocolumn
\renewcommand\twocolumn[1][]{%
    \oldtwocolumn[{#1}{
    \begin{center}
    \vspace{-2mm}
           \includegraphics[width=.99\textwidth]{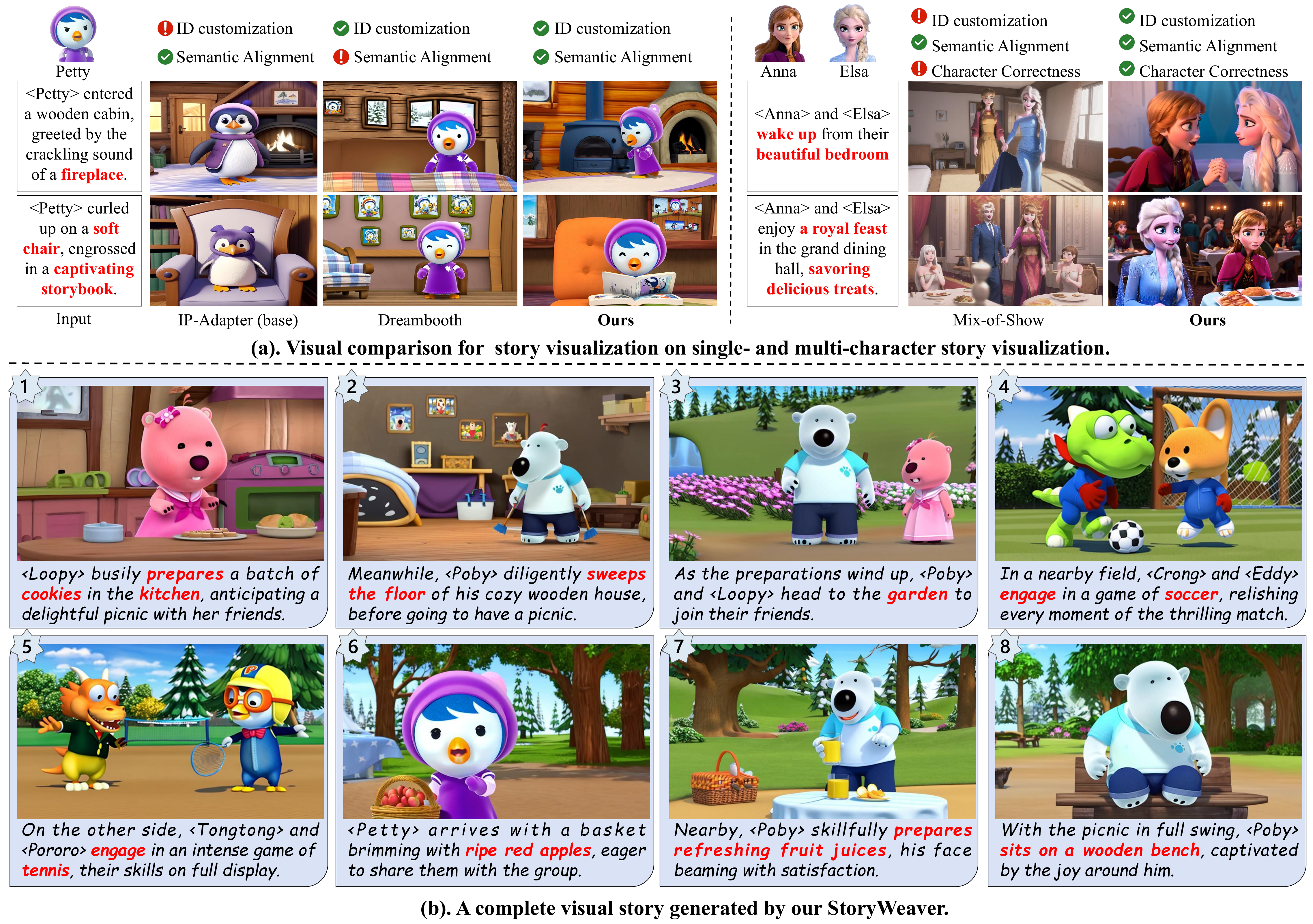}
           \vspace{-3mm}
           \captionof{figure}{Our StoryWeaver can achieve high-quality story visualization based on the given characters within a unified model.}
           \label{fig:FIG1}
        \end{center}
    }]
}

\begin{document}
\maketitle

\input{section/0_abs}
\input{section/1_intro}
\input{section/2_rela}
\input{section/3_method}
\input{section/4_exp}
\input{section/5_conclusion}

\input{section/6_ack}
\bigskip

\bibliography{aaai25}
\input{section/x_appendix}

\end{document}

%% file: section/0_abs.tex

\begin{abstract}
Story visualization has gained increasing attention in artificial intelligence. However, existing methods still struggle with maintaining a balance between character identity preservation and text-semantics alignment, largely due to a lack of detailed semantic modeling of the story scene. To tackle this challenge, we propose a novel knowledge graph, namely Character Graph (\textbf{CG}), which comprehensively represents various story-related knowledge, including the characters, the attributes related to characters, and the relationship between characters.
We then introduce StoryWeaver, an image generator that achieve Customization via Character Graph (\textbf{C-CG}), capable of consistent story visualization with rich text semantics. To further improve the multi-character generation performance, we incorporate knowledge-enhanced spatial guidance (\textbf{KE-SG}) into StoryWeaver to precisely inject character semantics into generation. To validate the effectiveness of our proposed method, extensive experiments are conducted using a new benchmark called TBC-Bench. The experiments confirm that our StoryWeaver excels not only in creating vivid visual story plots but also in accurately conveying character identities across various scenarios with considerable storage efficiency, \emph{e.g.}, achieving an average increase of +9.03\% DINO-I and +13.44\% CLIP-T. Furthermore, ablation experiments are conducted to verify the superiority of the proposed module. Codes and datasets are released at https://github.com/Aria-Zhangjl/StoryWeaver.
\end{abstract}

%% file: section/1_intro.tex
\section{Introduction}

Story Visualization is an emerging task in artificial intelligence with wide-ranging applications in education and entertainment, \emph{e.g.}, comic books creation and movie production ~\cite{li2019storygan, maharana2022storydall, zhou2024storydiffusion,cheng2024autostudio}. Given a textual narrative and portrait images of characters, the task of story visualization is to generate a series of images visually represent the story. Therefore, the main obstacle in this task is to customize the given characters faithfully and synthesize semantically diverse images that align well with the prompt along the whole storylines~\cite{gong2023talecrafter,su2023make,song2020character,chen2022character,cheng2024autostudio}.


Recent diffusion methods for character-consistent image generation ~\cite{gal2022image,ruiz2023dreambooth,ye2023ip,li2024blip,kumari2023multi,han2023svdiff} can be broadly categorized into two types, \emph{i.e.}, adapter-based and customization-based. Adapter-based methods ~\cite{ye2023ip,wang2024instantid, liu2024intelligent} introduce image-conditioned side network within diffusion models to provide visual guidance. For example, IP-Adapter ~\cite{ye2023ip} deploy an image adapter to extract feature from input images, while StoryGEN ~\cite{liu2024intelligent} incorporate a context module to provide visual context from previous frames. However, these methods struggle with detailed identity extraction and precise customization of characters. As depicted in Fig.~\ref{fig:FIG1}(a), IP-Adapter captures the coarse semantics of \emph{Petty}, \emph{e.g.}, a penguin wearing purple, but overlook finer details of the appearance. 


On another hand, customization-based methods ~\cite{ruiz2023dreambooth,hu2021lora,gal2022image,han2023svdiff} conceptualize the character on a set of customized images, achieving better identity preservation. However, as they are trained in an entangled way, \emph{i.e.}, rely on few text tokens to capture coarse-grained semantics, these methods are over-fit and hard to respond to text instructions, also illustrated in Fig.~\ref{fig:FIG1}(a). Furthermore, their per-concept optimization necessitates a distinct model for each character, resulting in a significant demand for storage resources.

Unlike previous works, we are keen to a unified framework with fine-grained modeling for comprehensive multi-character customization to achieve high-fidelity identity preservation and precise text alignments. 
Inspired by ERNIE~\cite{zhang2019ernie}, which enhance the language representation via external knowledge, 
we argue that finer-grained details within each story can be effectively captured with enhanced semantic-rich knowledge. 
Then we propose a novel Character-Graph (\textbf{CG}) to encode fine-grained semantics about the story world, including the given characters, their detailed attributes, and their relations. Characters are presented as object nodes in \textbf{CG} with multiple attribute node attached, and their relations serve as the edge to connect all objects together. These components collectively define the essence of each story scene. Then each visual story scene can be detailed into text captions through \textbf{CG}. By Customizing via Character-Graph enhanced scene-caption pairs (\textbf{C-CG}), our proposed model, namely StoryWeaver, is capable of capturing crucial semantics from the story scene, thereby dramatically improve the consistency and alignment on both identity and semantics.

However, multi-character story visualization is still intractable, which suffers from identity blending. This problem stems from incorrect attention distribution within diffusion model, where specific character knowledge will impact unrelated regions without spatial constraints. Existing methods ~\cite{gu2024mix,hu2021lora} adopt a regionally controllable sampling method to address this problem. As these methods necessitate additional spatial inputs, \emph{e.g.}, keypose image and layout, to strictly determine region assignment for different characters, they often encounter conflicts in layout and degradation in identity representation. As shown in the right side of Fig.~\ref{fig:FIG1}(a), identity variations among characters ``\emph{Anna}'' and ``\emph{Elsa}'' are evident across frames, and the model fails to generate accurate semantics of ``\emph{savor treats}''.

To remedy these issues, we propose Knowledge-Enhanced Spatial Guidance  (\textbf{KE-SG}) within the attention mechanism as external knowledge for precise multi-character customization without quality compromise. Specifically, we introduce a knowledge encoder to extract the features of different characters, then refine the initial position prior by the extracted character knowledge to modify the incorrect cross-attention map. Through \textbf{KE-SG}, character knowledge from \textbf{CG} is accurately attend to the corresponding regions in the story scene, ensuring precise identity representation and coherent text semantics for multi-character visual storytelling tasks. As shown in Fig.~\ref{fig:FIG1}(b), StoryWeaver achieves vivid visual plot generation contains encompassing both single and multi-character interactions.

Moreover, we introduce a new benchmark termed TBC-Bench to train our StoryWeaver for both single- and multi-character story visualization, and compare it with a set of state-of-the-art (SOTA) methods~\cite{ye2023ip,liu2024intelligent,ruiz2023dreambooth,hu2021lora,gu2024mix,yang2024lora}. The experimental results demonstrate that our StoryWeaver excels not only in character identity preservation, \emph{e.g.,} achieving an average increase of +9.03\% DINO-I on single-character customization, but also the visual quality in various tasks with an average increase of +18.45\% CLIP-T and + 19.11\% Character F1 for multi-character generation. 

To sum up, the contributions of this work are three-fold:
\begin{itemize}
    \item We propose a novel Character Graph to structurally represent semantic-rich knowledge within each story world and a novel StoryWeaver enhanced by the structured knowledge to achieve high-quality visual storytelling.
    \item We introduce a novel knowledge-enhanced spatial guidance (\textbf{KE-SG}) for precise cross-attention assignment to address identity blending, which improves the performance of multi-character generation.
    \item Our StoryWeaver outperforms a set of compared methods on the newly proposed TBC-Bench in terms of character identity preservation, correct complex scene generation and text semantics alignment.
\end{itemize}

%% file: section/2_rela.tex
\section{Related Work}

\subsection{Story Visualization}
The task of Story visualization to generate image series aligning with multi-sentence paragraphs while maintaining global semantic consistency throughout the storyline. Earlier works have either adapted GAN for this task~\cite{li2019storygan,song2020character,li2022clustering,li2020improved,szHucs2022modular}, followed by Transformers-based methods ~\cite{chen2022character,maharana2022storydall} which leverage the long-range dependence properties to enhance semantic coherence. 

Recent studies ~\cite{feng2023improved,rahman2023make,su2023make,liu2024intelligent,gong2023talecrafter,zhou2024storydiffusion,cheng2024autostudio,tewel2024training,Omri2024the} have explored diffusion model~\cite{rombach2022high} to achieve consistent image generation, especially for open-ended visual storytelling task\cite{zhou2024storydiffusion,liu2024intelligent,cheng2024autostudio}. For example, StoryGEN ~\cite{liu2024intelligent} trains a visual language context module to extract information from previous-turn images while Storydiffusion ~\cite{zhou2024storydiffusion} proposes Consistent Self-Attention within a batch. 
\begin{figure*}[!t]
    \centering
    \vspace{-6mm}
    \includegraphics[width=1.\linewidth]{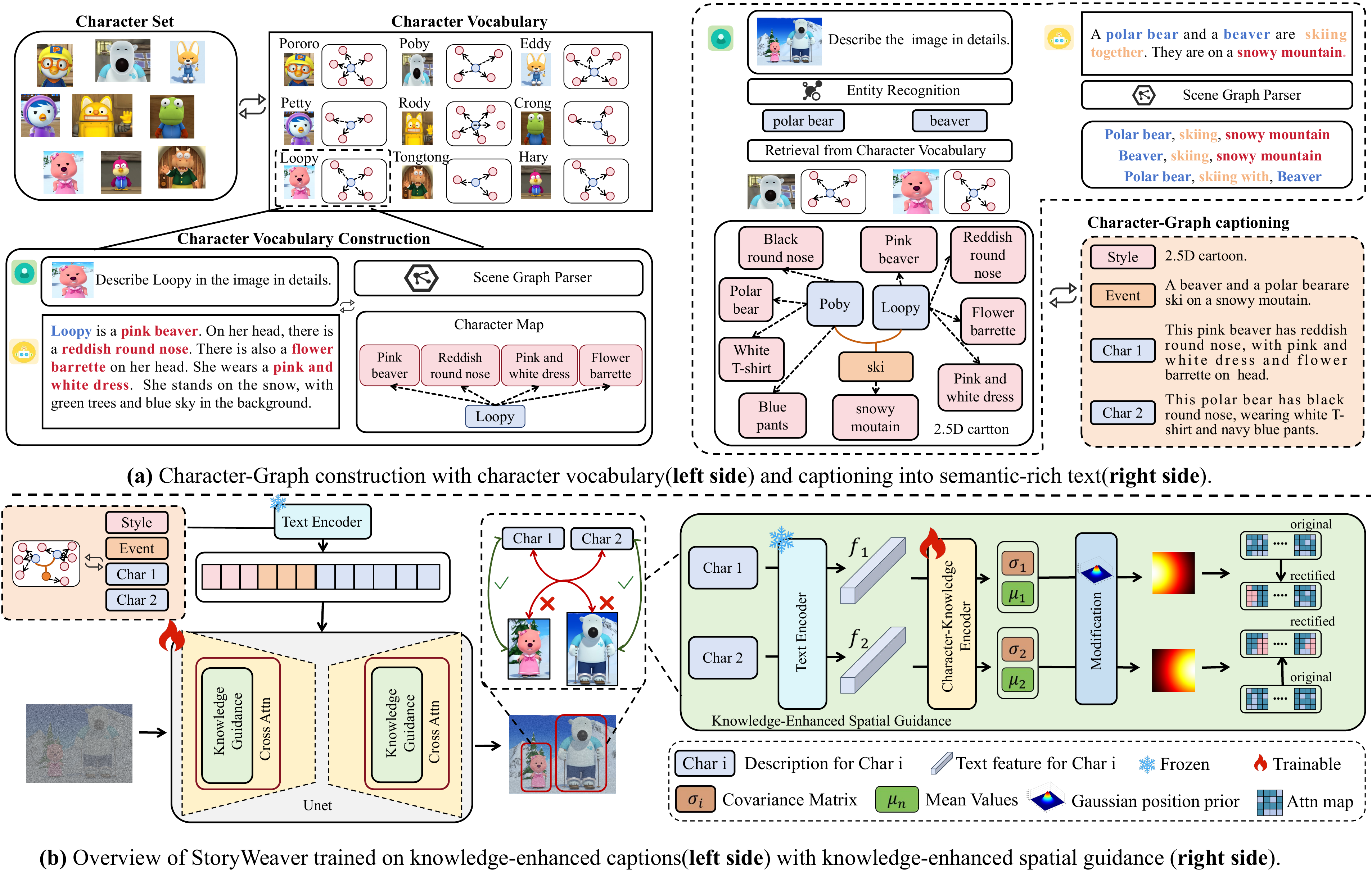}
    \vspace{-6mm}
    \caption{The overall framework of StoryWeaver. \textbf{(a).} We propose Character-Graph to represent semantic-rich knowledge within the story world. \textbf{(b).} We enhanced the StoryWeaver  with proposed spatial guidance for further improving the performance of mult-character generations.}
    \vspace{-6mm}
    \label{fig:overview}
\end{figure*}
\subsection{Image Customization}
Image Customization aims to synthesize specific subjects align with given textual contexts. Single-concept customization methods either learn new ``word'' embeddings from small image sets of customized subjects ~\cite{gal2022image,voynov2023p,kumari2023multi,dong2022dreamartist} or train diffusion models with additional modules to encode visual guidance~\cite{ruiz2023dreambooth,wei2023elite,chen2024subject,jia2023taming,li2024blip,ma2023unified,yuan2023customnet,hu2021lora}. 

While significant progress has been made, multi-concept customization remains challenging ~\cite{han2023svdiff,yang2024lora,xiao2023fastcomposer}. To address multi-subject identity blending issue, existing methods use additional loss on large multi-subject dataset or rely on extra optimization efforts to merge multiple models ~\cite{xiao2023fastcomposer,shentu2024attencraft,zhang2024attention,gu2024mix,xie2023boxdiff,yang2024lora}. Some methods ~\citep{gu2024mix, xie2023boxdiff, zhou2024migc} use spatial control inputs like layout and bounding box annotation to separate, still facing challenges in natural interaction synthesis and per-character identity preservation. 

%% file: section/3_method.tex
\section{Method}

\subsection{Overview}
In this paper, 
we aim to achieve fine-grained story world simulation by a unified model for story visualization, as depicted in Fig.~\ref{fig:FIG1}(b).
The overview of Storyweaver is shown in Fig.~\ref{fig:overview}.
We first propose the Character-Graph, a novel representation of the semantic-rich knowledge within this particular story world, to enhance consistency generation in diffusion model (Fig.~\ref{fig:overview}(a)). Subsequently, we employ spatial guidance to effectively incorporate the Character-Graph knowledge for precise multi-character generation (Fig.~\ref{fig:overview}(b). In the following sections, we will introduce customization via the Character-Graph (\textbf{C-CG}) and then detail the knowledge-enhanced spatial guidance (\textbf{KE-SG}).

\subsection{Customization via the Character-Graph}

Existing methods \cite{ruiz2023dreambooth, hu2021lora, gong2023talecrafter, rahman2023make} use simple tokens for character customization across a few samples to achieve consistent image generation. Inspired by ERNIE \cite{zhang2019ernie}, which employs an external knowledge map to enhance token representation in language model(\textbf{LM}), we integrate a novel Character-Graph to enhance the representation of story scenes, thereby improving character identity preservation and semantic modeling.

\subsubsection{Character-Graph Construction.}
Detailed semantics, including objects, their attributes, and relationships, are crucial to the understanding of visual scenes ~\cite{johnson2015image}.  
In the story world, characters act as pivotal \textbf{objects} that form the most important part of the world. \textbf{Attributes} linked to characters are also crucial, as they vividly depict each character's appearance. Interactions among characters unveil the most intricate dynamics within the story world, representing the unfolding \textbf{events} that propel the story forward. As illustrated in Fig.~\ref{fig:overview}(a), Character-Graph in a story world can be formulated as $G=<O, E, A>$, where $O$ represents the character sets, $E$ denotes the set of events, and $A$ refers to the set of attribute nodes associated with $O$ and $E$.

To construct $G$, we begin by creating a character vocabulary for the given character set. We collect frontal image $I_i$ for each character, as depicted in Fig.~\ref{fig:overview}(a). Then, we use a vision-language model (\textbf{VLM}) to extract a detailed caption with rich semantic from the image, formulated as:
\begin{equation}
C_i=V_{cap}(Instruct_c, I_i), i \in [1, N_c]\, ,
\end{equation}
where $C_i$ is the caption of $I_i$ obtained by prompting the VLM model denoted as $V_{cap}$ with an instruction $Instruct_c$, and $N_c$ is the number of characters in the character set.

However, while such a detailed caption contains rich attributes of the character, it also contains unrelated semantics that are useless for customization. Therefore, we further propose a parsing method to extract detailed semantics related only to the character $O_i$ by:
\begin{small}
\begin{equation}
\sum Map<O_i, A_i^k> =  (SG_i^{(A)} | O_i) =   (Parser(C_i) | O_i).
\label{equ:char-map}
\end{equation}
\end{small}%
Here $SG_i^{(A)}$ is the scene graph for image $I_i$ obtained by a Scene Graph Parser ~\cite{wu2019unified} based on $C_i$, denoted as $Parser$. $<O_i, A_i^k>$ is the character map constructed by the character $O_i$ itself with the related $k_{th}$ attribute denoted as $A_i^k$. As depicted in Fig. \ref{fig:overview}(a), \emph{``Loopy"} is described by detailed attributes, \emph{e.g., ``reddish round nose"}. 

Besides, \textbf{events}, \emph{e.g.},  ``hugging" and ``kissing", are used to illustrate the connections among or within characters. Similarly, for a given story scene $\mathcal{F}$ containing character $i$ and character $j$, we first extract the rich semantic caption $\mathcal{F}_c$ via a \textbf{VLM} model, and then a Scene Graph Parser is adopted to extract the event-related semantic: 
\begin{small}
\begin{equation}
R(O_i, O_j) = (SG_i^{(R)} | (O_i, O_j)) = (Parser(\mathcal{F}_c))_{i, j} ,
\end{equation}
\end{small}%
where $\mathcal{F}_c$ is the original scene caption, and $R(O_i, O_j)$ denotes the events between $O_i$ and $O_j$\footnote{\emph{$O_i$} and \emph{$O_j$} may refer to the same character}. 
Then, Character-Graph that describes the extensive character and event knowledge within the story's scenes can be formulated as: 
\begin{small}
 \begin{equation}
G(O, E) = \{ \sum_{i, k} Map<O_i, A_i^k>, \sum_{i, j} R(O_i, O_j)\}.
\end{equation}   
\end{small}%

\subsubsection{Scene Caption via Character-Graph.}
Based on Character-Graph, we create a detailed description for each story scene, encompassing the fine-grained attributes of characters and the interactions between them. With this structured and semantic-rich knowledge-enhanced captions, the model can achieve better consistencies of characters and semantic alignment of events.
\begin{figure*}[!t]
    \centering
    \includegraphics[width=1.\textwidth,height=.3\textwidth]{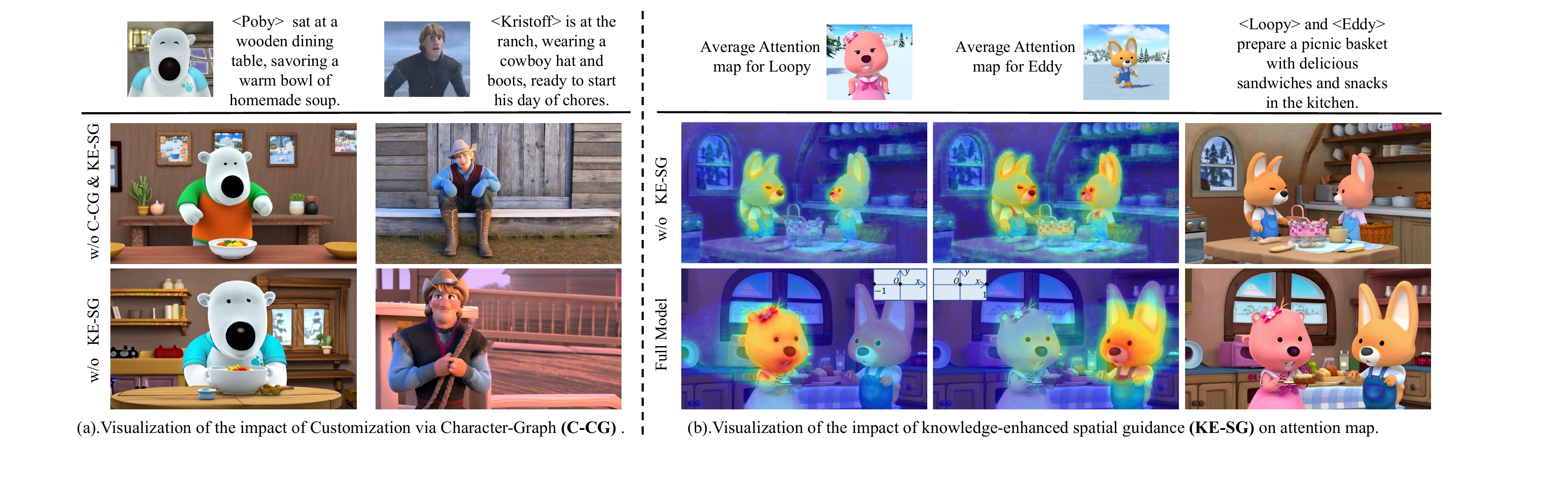}
    \vspace{-5mm}
    \caption{Visual examples for the impact of Customization via Character-Graph (\textbf{C-CG}) and Knowledge-Enhanced Spatial Guidance (\textbf{KE-SG}). \textbf{(a).}Without \textbf{C-CG}, the generator struggles to capture finer-grained details of character. \textbf{(b).}Without \textbf{KE-SG}, the generator tends to allocate attention uniformly across all regions, resulting in identity blending. }
    \label{fig:ablation}
\vspace{-5mm}
\end{figure*}%

For each story scene $\mathcal{F}$, we first employ $V_{cap}$ to generate a detailed description prompted by an instruction $Instruct_e$, \emph{e.g.},  ``A polar bear and a beaver are skiing on a snowy mountain." 
 Subsequently, we extract the characters and their relations by Scene Graph Parser:
\begin{small}
\begin{equation}
\begin{aligned}
\sum_j R_{j, *}, char_j  = Parser( V_{cap}(Instruct_e, \mathcal{F})),
\end{aligned}
\end{equation}
\end{small}%
where $char_j$ is the $j_{th}$ character coarse label, \emph{e.g.}, ``bear" and ``beaver" for the given example, and $R_{j, *}$ refers to the relationship between $char_j$ and object $O_*$.

Then we lookup the character vocabulary in $G$ to identify the exact characters with the linked attributes formulated as:
\begin{small}
\begin{equation}
\begin{aligned}
O_{char}^j  =  \mathop{\arg\max}\limits_{O_i} Sim(char_j, O_i),\\
A_{char}^{j}=  O_{char}^j \otimes \sum Map<O_i, A_i^k>.
\end{aligned}
\end{equation}
\end{small}%
where $O_{char}^j$ is the matched character of $char_j$ by a matching function $Sim$, $A_{char}^{j}$ denotes the linked attributes associated with $O_{char}^j$ and $\otimes$ refers to the lookup operation. 



We then serialize the structured knowledge graph into a scene caption for customization. First, we combine $O_{char}^{j}$ 
 with the related attributes to generate an appearance description $W_c^{j}$ for $char_j$ by:
\begin{small}
\begin{equation}
W_c^{j} = O_{char}^j \oplus A_{char}^j,
\end{equation}
\end{small}%
where $\oplus$ denotes the union operation. Next, we use the relationships between objects to describe the events in the scene by $W_e$, where:
\begin{small}
\begin{equation}
W_e = \sum_{j, j^{\prime}} O_{char}^j \oplus R_{{j, j^{\prime}}} \oplus O_{char}^{j^{\prime}}.
\end{equation}
\end{small}%
 Furthermore, we employ a descriptive sentence to characterize the style of all scenes within the story's world, denoted as $W_s$. Then, the complete story scene caption can be formulated as $T_{g} = [W_{s}, W_{e}, \sum_j^N W_{c}^{j} ] $ where $N$ is the number of character(s) appear in that scene.

In this case, given scene image frame $\mathcal{F}$ and the obtained $T_g$, the objective of the diffusion model is:
\begin{equation}
       \mathbb{E}_{f\sim E(\mathcal{F}),T_{g},\epsilon\sim\mathcal{N}(0,1),t}\Big[\|\epsilon-\epsilon_\theta(f_t,t,\tau(T_{g}))\|_2^2\Big],
       \label{equ:objective}
\end{equation}
where $\epsilon$ is a random noise, $t$ is the sampled timestep, $E$ is the VAE encoder and $\tau$ is the text encoder. We name the model enhanced by C-CG as StoryWeaver because it seamlessly ``weave'' all elements within $G$ in a unified model and achieve improved customization for story visualization.

\subsection{Knowledge-Enhanced Spatial Guidance for Multi-Character Generation}



 Akin to previous studies \cite{xiao2023fastcomposer, yang2024lora,han2023svdiff}, StoryWeaver faces challenges with identity blending in multi-character generation. 

\noindent \textbf{Attention Blending.} In our case, the cross-attention mechanism in the diffusion model that used to update image feature $f_\mathcal{F}$ can be formulated as:
\begin{small}
\begin{equation}
\begin{aligned}
Attn=\mathcal{M}\cdot V =\text{Softmax}\Big(\frac{(W_qf_\mathcal{F}) (W_kf_T)^T}{\sqrt{d}}\Big)\cdot (W_vf_T),%
\end{aligned}
\label{equ:cross-attn}
\end{equation}
\end{small}
where $\mathcal{M}$ denotes the attention map between text feature $f_T$ and image feature $f_\mathcal{F}$. And $W_q$, $W_k$ and $W_v$ are learnable projection matrices, and $d$ is the latent projection dimension. 


Given $T_{g} = [W_{s}, W_{e}, \sum_j^N W_{c}^{j}] =w[1:T]$, the cell $\mathcal{M}_{x, y}^{i}$ represents the correlations between the word $w[i]$ and the image at pixel ($x$, $y$). 
Considering a situation where $w[i] \in W_c^{j}$ is part of the description of $O_{char}^j$ with extremely high $\mathcal{M}_{x, y}^{i}$, where ($x$, $y$) should correspond to another character $O_{char}^{j^{\prime}}$, the incorrect knowledge guidance of $\mathcal{M}_{x, y}^{i}$ can be highly detrimental to generation of consistent characters.
This is evidenced by the visualization in Fig.\ref{fig:ablation}(b), where the average attention maps for Eddy's description contribute equally to both regions, leading to identity blending with duplicate character generation.
\begin{table*}[!t]
\small{
\renewcommand{\arraystretch}{1.09}
\setlength{\tabcolsep}{0.1mm}{
\vspace{-1mm}
\begin{tabular}{cccccccccc}
\hline
\multirow{2}{*}{\textbf{Task}} &\multirow{2}{*}{\textbf{Type}}   &\multirow{2}{*}{\textbf{Method}} &  \multirow{2}{*}{\textbf{\# Params/DB($\downarrow$)}} & 
\multicolumn{3}{c}{\textbf{Pororo}} & 
\multicolumn{3}{c}{\textbf{Frozen}}  \\
\cline{5-7} \cline{8-10}
& & & & \textbf{DINO-I($\uparrow$)} & \textbf{CLIP-I($\uparrow$)} &\textbf{CLIP-T($\uparrow$)} &   \textbf{DINO-I($\uparrow$)} & \textbf{CLIP-I($\uparrow$)} & \textbf{CLIP-T($\uparrow$)} \\
\hline
\multirow{6}{*}{Sin-Char} & \multirow{3}{*}{Adapter-based}& StoryGEN &  1064 M & 52.92&76.03&26.98& 46.67&72.61&28.05\\
& & IP-Adapter(base) &  1038 M &  48.85 &76.66&29.98&44.15&78.42&31.69\\
& & IP-Adapter(plus) & 1063 M  & {\color{gray}64.36}&{\color{gray}81.64}&{\color{gray}24.88}& {\color{gray}60.87}&{\color{gray}84.52}&{\color{gray}27.15} \\

\cline{2-10}
&\multirow{3}{*}{Customization-based}& LORA &  1024 M   &54.13 &75.19&28.53 & 49.02&82.77&29.18\\
& & Dreambooth & 7118 M & 61.85&78.86&26.74& 55.01&81.07&27.12 \\
& & \textbf{StoryWeaver(ours)} &  1017 M &\textbf{64.96}&\textbf{82.65}&\textbf{33.26} &\textbf{62.17}&\textbf{85.24}&\textbf{36.74}\\
\hline
\hline
\multirow{2}{*}{\textbf{Task}} &\multirow{2}{*}{\textbf{Type}} &  \multirow{2}{*}{\textbf{Method}} &  \multirow{2}{*}{\textbf{\# Params/DB($\downarrow$)}} & 
\multicolumn{3}{c}{\textbf{Pororo}} & 
\multicolumn{3}{c}{\textbf{Frozen}}  \\
\cline{5-7} \cline{8-10}
& & & & \textbf{CLIP-T($\uparrow$)} & \textbf{F-Acc($\uparrow$)} &\textbf{C-F1($\uparrow$)} &  \textbf{CLIP-T($\uparrow$)} & \textbf{F-Acc($\uparrow$)} &\textbf{C-F1($\uparrow$)} \\
\hline
\multirow{4}{*}{Multi-Char}&Adapter-based & StoryGEN &  1064 M &  27.27 & 19.55 &27.17 & 28.91 & 12.31&21.79\\
\cline{2-10}
& \multirow{3}{*}{Customization-based}& Mix-of-Show & 1164 M & 27.20 & 30.23 &44.03& 30.71 & 18.90&30.62\\
& & LoRA-Composer &  1425 M  & 27.86 & 27.04 & 47.36 & 28.88 & 27.69&39.72 \\
& & \textbf{StoryWeaver(ours)} & 1017 M   &\textbf{34.30}&\textbf{40.45}&\textbf{59.72}&\textbf{34.94}&\textbf{34.51}&\textbf{44.53}\\
\hline
\end{tabular}
}
\vspace{-2mm}
\caption{Quantitative comparisons on the single- and multi-character generation with existing methods. Our StoryWeaver obviously merits in semantic alignments with high identity customization compared to existing methods. 
\vspace{-1mm}
}
\label{tab:qualitative}
}
\vspace{-4mm}
\end{table*}

\noindent \textbf{Knowledge-Enhanced Spatial Guidance.}
Since identity blending arises from an imperfect attention map, we propose a knowledge-enhanced spatial guidance (KE-SG) to modify the attention maps for precise multi-character generation.

Specifically, 
we first assign a position prior for character $O_{char}^{j}$ as an external knowledge, which follows the Gaussian Distribution and can be  formulated as:
\begin{equation}
    p_j(x,y) = \frac{1}{2 \pi \sqrt{|\Sigma|} } e^{ -\frac{1}{2}(x-\mu_x, y-\mu_y)\Sigma^{-1}(x-\mu_x, y-\mu_y)^T},
\label{equ:position_prior}
\end{equation}
where $\mu_x$ and $\mu_y$ represent the mean values in the horizontal and vertical directions, and $\Sigma$ denotes the covariance matrix. 
$p_j(x,y)$ denotes the probability of pixel $(x,y)$ belongs to $O_{char}^{j}$.
In practice, we assume that characters appearing in the same scene are evenly distributed horizontally. Therefore, we initially set $\mu_y = 0$ and $\Sigma=I$ for all characters while $\mu_x = -1 + j \times \frac{2}{n-1}$ for character $O_{char}^{j}$, where $n$ is the total number of characters in the scene\footnote{The coordinate of the position prior is illustrated in Fig.\ref{fig:ablation}(b)}.

As characters vary in size, we leverage the knowledge from $W_c^{j}$ to precisely modify the $p_j(x,y)$. Given the corresponding appearance description $W_c^{j}$, we propose a knowledge encoder $\mathcal{E}$ to extract the spatial semantics of the character, formulated as:
\begin{equation}
    \mathcal{E}(W_c^{j}) \to \{(\Delta \mu_x, \Delta \mu_y), \gamma\},
\label{equ:mask_predicot}
\end{equation}
where 
$(\Delta \mu_x, \Delta \mu_y)$ represent the offset of mean and $\gamma$ represents the scale of corvariance. Then, the knowledge-enhanced position prior can be obtained by:
\begin{equation}
\begin{aligned}
       qi
\label{equ:position_prior_text}
\end{aligned}
\end{equation}
where $t_j$ denotes the text feature for $W_c^{j}$.

We sample the character-aware spatial guidance $\mathcal{P}^j$ for $O_{char}^{j}$ by Eq.\ref{equ:position_prior_text}. Then, the character-related region on $\mathcal{M}^i$ is enhanced along with the unrelated region decreased by:
\begin{equation}
    [\mathcal{M}_{t}^i]_{(x,y)}=\left\{
\begin{aligned}
[\mathcal{M}_{t}^i]_{(x,y)} + s, & \text{if} \ \ [\mathcal{P}^j]_{(x,y)} \ge \beta, \\
[\mathcal{M}_{t}^i]_{(x,y)} - s, & \text{if} \ \ [\mathcal{P}^j]_{(x,y)} < \beta.
\end{aligned}
\right.
\end{equation}%
where $[\mathcal{M}_{t}^i]$ refers to the cross-attention map for token $w[i] \in W_c^j$ at $t$ timestep and $(x,y)$ is the pixel coordinate, $\beta$ is the threshold determine whether image pixel $(x,y)$ is related with $O_{char}^j$. The time-aware guidance scale $s$ is:
\begin{equation}
    s=s(t)=\alpha \cdot (ln(t+1)+1),
    \label{equ:attention_strength}
\end{equation}
where $\alpha$ is the guidance strength and $t$ is the timesteps. $s$ applies stronger knowledge guidance in the initial steps during the early noisy stages, gradually tapering off the guidance strength to prevent quality degradation. 

As shown in Fig.\ref{fig:ablation}(b), StoryWeaver achieve correct multi-character generation well-matched the given text instruction.

%% file: section/4_exp.tex
\section{Experiments}

\subsection{Experimental Settings}
\subsubsection{Dataset Construction} 
Existing datasets \cite{li2019storygan,gupta2018imagine} for story visualization suffers from low resolution and simple caption annotations.
We focus on two popular cartoon series, \emph{i.e.}, \emph{Pororo the Little Penguin} and \emph{Frozen} and construct a dataset featuring multiple cartoon characters. 
Due to resource constraints, we selected around 10 images per character depicting various events, including over 30 images with multiple characters to capture complex interactions.
We also established an evaluation benchmark termed \textbf{TBC-Bench}, including 5 single-character stories for each character and 10 multi-character stories in all.


\subsubsection{Implement Details} 
We employed Stable-Diffusion v1-5 and implement character knowledge encoder $\mathcal{E}$ as MLP. The whole model is trained with a learning rate of $7 \times 10^{-6}$ and a batch size of 4. For characters from \emph{Pororo}, we set $\alpha=2.5$, whereas $\alpha=1$ for \emph{Frozen} and $\beta=0.85 \times \Vert \mathcal{P}^j \Vert_{max}$. During inference, we employed the LMSDiscrete Scheduler with 100 sampling steps and the text guidance scale is 5.0.

\subsubsection{Evaluation Metric}
Following \cite{zhou2024storydiffusion, cheng2024autostudio, gu2024mix}, we evaluate the methods from three aspects: (1) \textit{Identity Preservation}, \emph{i.e.}, calculating the similarity between generated image and GT character image by DINO \cite{oquab2023dinov2} (\textbf{DINO-I}) and CLIP \cite{radford2021learning} (\textbf{CLIP-I}), (2)\textit{Semantic alignment}, \emph{i.e.}, calculating the text-image similarity of the generated image and the text description by CLIP (\textbf{CLIP-T}) , (3)\textit{Character Integrity}, \emph{i.e.}, Character F1 (\textbf{C-F1}) that measuring the percentage of characters in a generated image that exactly match the story input, and Frame Accuracy (\textbf{F-Acc}) that measures whether all characters in a story are present in the generated image. We use a pre-trained DINO as the classifier, where a similarity score exceeding 0.5 between the renderings and the ground truth character image indicates a correct sample. 

\subsection{Single Character Customization}


\subsubsection{Quantitative Results}
\begin{figure*}[!t]
    \centering
    \vspace{-5mm}
    \includegraphics[width=1.\textwidth]{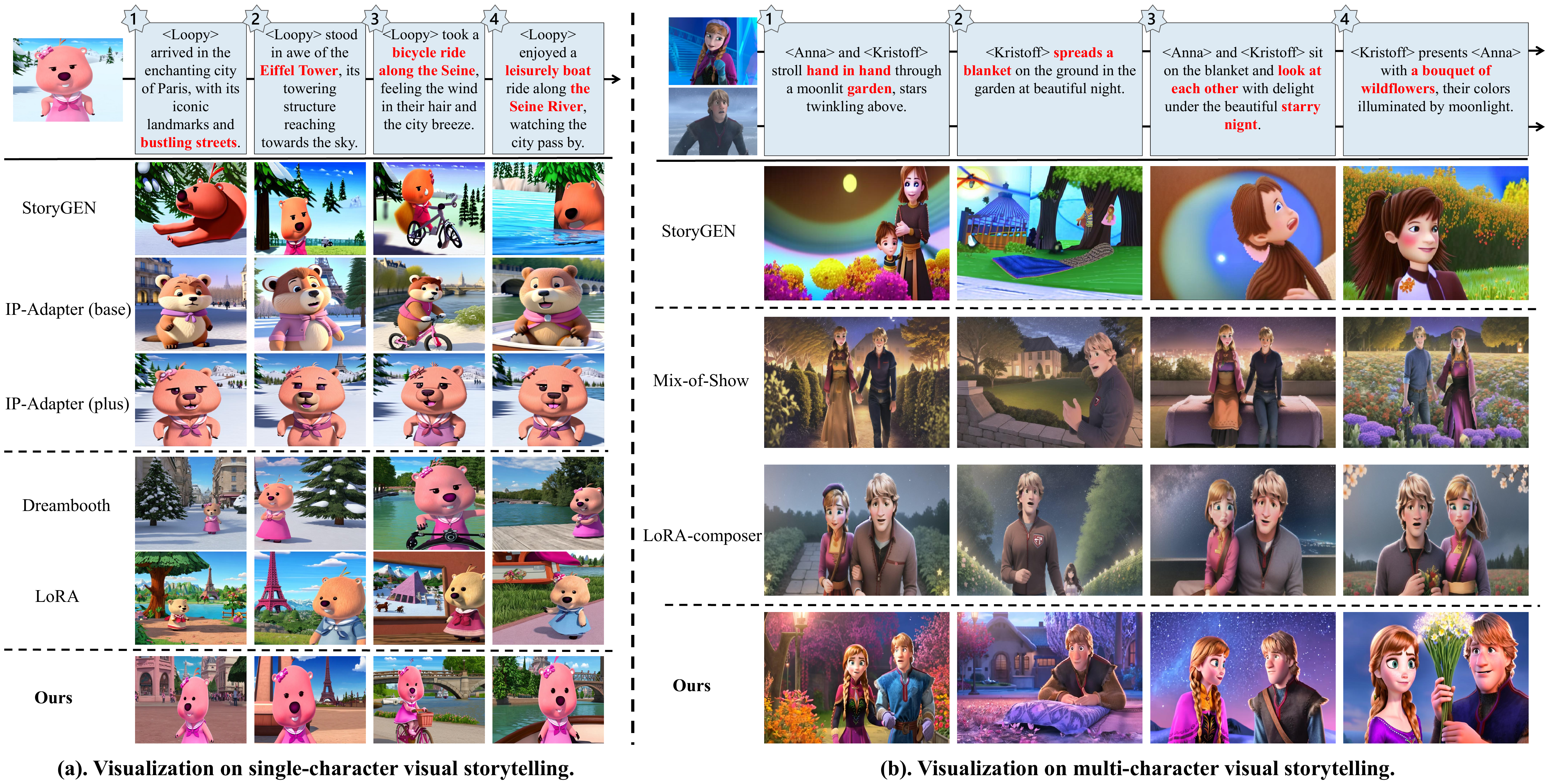}
    \vspace{-6.5mm}
    \caption{The visual comparisons of different methods on single and multi-character visual storytelling.
    Our StoryWeaver excels in  character identity customization and well-matched semantic alignment.
    }
    \label{fig:single-Loopy}
\vspace{-3mm}
\end{figure*}

From Tab.\ref{tab:qualitative}, it is evident that IP-Adapter-plus exhibits the highest character consistency with the poorest text semantic alignment. As shown in Fig.~\ref{fig:single-Loopy}(a), its generations are minor editions of the reference image without text semantic alignments.
Secondly, the two other customization-based methods achieve better character consistency than adapter-based ones. However, the declination on CLIP-T indicates that such improvement comes at the cost of semantic alignment. 
Finally, StoryWeaver surpasses all other methods, significantly improving character consistency by +13.02\% on DINO-I and text semantic alignment by +15.94\% on CLIP-T for \emph{Frozen}, demonstrating the superiority of our proposed Character Graph modeling.

\subsubsection{Visualization Results}
As shown in Fig.\ref{fig:single-Loopy}(a), StoryGEN and IP-Adapter (base) struggle to customize characters with intricate details. The generated images bear minimal resemblance to ``\emph{Loopy}''. 
Other customization-based methods like Dreambooth and LoRA face challenges in balancing faithful character customization with semantic alignment. 
Conversely, with semantic-rich knowledge enhanced, StoryWeaver excels in crafting high-quality images with faithful identity preservation, aligning well with the  prompts.

\subsection{Multi Character Customization}
\subsubsection{Quantitative Results}

From Tab.\ref{tab:qualitative}, we can first find that existing methods all fails to achieve semantic-aligned generation. However, our proposed Storyweaver significantly enhances text semantic alignments (+23.12\% on \emph{Pororo} and +13.77\% on \emph{Frozen}) while also notably improving character consistency(+26.10\% in C-F1 on \emph{Pororo} and + 12.11\% on \emph{Frozen}). Experimental results demonstrate that our unified modeling and \textbf{KE-SG} substantially enhances the efficacy and stability of multi-character generation.

\begin{table}[!t]
\centering
\footnotesize{
\renewcommand{\arraystretch}{1.05}
\setlength{\tabcolsep}{5.5mm}{
\begin{tabular}{c|ccc}
\hline
\multirow{2}{*}{\textbf{Methods}} 
& \multicolumn{3}{c}{\textbf{Single Character Generation}} \\
& \textbf{T-A} & \textbf{I-A} & \textbf{V-Q}\\
\hline
StoryGEN & 1.32 & 1.31 & 1.11  \\
IP-Adapter (base) & 3.01 &2.10  &2.61 \\
IP-Adapter (plus) & 2.22 &3.73  & 3.29  \\
\hline
LoRA & 2.71 & 1.94 & 2.56   \\
Dreambooth & 2.47 & 3.19 & 2.56 \\
\hline
Ours & \textbf{4.05} & \textbf{4.13} &\textbf{4.35} \\
\hline
\hline
\multirow{2}{*}{\textbf{Methods}}& \multicolumn{3}{c}{\textbf{Multi Character Generation}} \\
& \textbf{T-A} & \textbf{I-A} & \textbf{V-Q}\\
\hline
StoryGEN & 1.36 &1.58 & 1.43    \\
\hline
Mix-of-Show & 2.74 & 3.47 &3.44  \\
LoRA-Composer & 3.25 &3.50  & 4.03  \\
\hline
Ours & \textbf{4.25} &\textbf{4.31}  & \textbf{4.32}  \\
\hline
\end{tabular}
}
\vspace{-2mm}
\caption{User Study. A higher score indicates better performance in terms of the corresponding metric.
}
\label{tab:user_study}
}
\vspace{-4mm}
\end{table}
\subsubsection{Visualization Results}
As shown Fig.\ref{fig:single-Loopy}, StoryGEN struggles with responding the given prompt, mainly due to the absence of explicit learning for the character and event. For instance, neither ``\emph{Anna}'' nor ``\emph{Kristoff}'' are generated correctly in its renderings. Multi-concept tuning methods are much better in identity preservation. However, due to the strong input spatial constraints of keypose or layout, these two methods synthesizes unnatural interaction and barely achieve aligned text semantics, as evidenced by the depictions of ``\emph{starry night} '' and ``\emph{wildflowers}''. Conversely, StoryWeaver can well respond to the text semantics while retaining high character consistency.

\subsection{User Study}
Following \cite{su2023make}, we conducted a user study with 50 participants, evaluating existing methods across five single-character stories and five multi-character stories. Participants are prompted to rank the methods based on \emph{Text-alignment} (\textbf{T-A}), \emph{Image-alignment} (\textbf{I-A}) and \emph{Overall Visual Storytelling Quality} (\textbf{V-Q}), assigning scores from 5 to 1 for each ranking. The results in Tab.\ref{tab:user_study} highlight the superior performance of our StoryWeaver over others, affirming the effectiveness of our proposed StoryWeaver.

\begin{table}[!t]
\centering
\footnotesize{
\renewcommand{\arraystretch}{1.1}
\setlength{\tabcolsep}{0.4mm}{
\begin{tabular}{c|ccc|ccc}
\hline
\multirow{2}{*}{\textbf{Setting}}
& \multicolumn{3}{c|}{\textbf{Single Character}} & \multicolumn{3}{c}{\textbf{Multi Character}}\\

& DINO-I & CLIP-I & CLIP-T& CLIP-T & F-Acc &C-F1 \\
\hline

w/o (C-CG \& KE-SG) &39.57 & 68.11 &28.74 & 30.98 & 22.22& 48.26 \\
\hline
\textcolor{gray}{dreambooth} &\textcolor{gray}{61.85} & \textcolor{gray}{78.86} &\textcolor{gray}{26.74} & - & - &  - \\
w/o KE-SG &60.98 & 81.28 &32.17 & 33.41 & 35.73&  52.69 \\
\hline
StoryWeaver &\textbf{64.96} & \textbf{82.65 }&\textbf{33.26} & \textbf{34.30} & \textbf{40.45}& \textbf{59.72}\\
\hline
\end{tabular}
\vspace{-2mm}
\caption{Ablation study of on Pororo. \emph{\textbf{C-CG}} refers to customization via Character-Graph and \emph{\textbf{KE-SG}} refers to the knowledge-enhanced spatial guidance. 
}
\vspace{-5mm}
\label{tab:ablation}
}
}
\end{table}
\subsection{Ablation Studies}
In Tab.\ref{tab:ablation}, we conduct ablations on different designs of StoryWeaver. In comparison to the plain baseline, \emph{i.e.}, devoid of Customization via Character Graph (C-CG) and knowledge-enhanced spatial guidance (KE-SG), C-CG significantly enhances the fidelity of the character's identity. Additionally, KE-SG effectively addresses identity blending issues, leading to the highest Frame-Accuracy and Character F1 scores in the full model settings. These findings are also supported by the results presented in Fig.\ref{fig:ablation} as previously discussed.

%% file: section/5_conclusion.tex
\section{Conclusion}
In this paper we introduce StoryWeaver, a unified model with intricate characters customization for story visualization. We first propose a novel Character-Graph that encapsulate semantic-rich knowledge within the story world to enhance our StoryWeaver. Then, we introduce knowledge-enhanced spatial guidance to refine cross-attention maps for precise multi-character generation. Experiment results demonstrates that StoryWeaver achiever better hither-fidelity in identity customization and better semantic alignment than a set of single- and multi-customization methods.

%% file: section/6_ack.tex
\section{Acknowledgments}
This work was supported by National Key R\&D Program of China (No.2023YFB4502804), the National Science Fund for Distinguished Young Scholars (No.62025603), the National Natural Science Foundation of China (No. U22B2051, No. U21B2037, No. 62072389, No. 62302411), the Natural Science Foundation of Fujian Province of China (No.2021J06003), China Postdoctoral Science Foundation (No. 2023M732948), and partially sponsored by CCF-NetEase ThunderFire Innovation Research Funding (NO. CCF-Netease 202301).

%% file: section/x_appendix.tex
\newpage
\section{Appendix}
In this supplement, we begin by outlining the construction details of TBC-Bench and the implementation specifics of our proposed customization using Character-Graph. Following this, we introduce the comparison methods employed and detail their respective experimental setups. Subsequently, we offer a thorough explanation of the evaluation metrics used in the quantitative comparison. Finally, we present additional results to further demonstrate the effectiveness of our proposed StoryWeaver. The code and proposed TBC-Bench is released anonymously at \url{https://anonymous.4open.science/r/StoryWeaver-C8A4}.
\subsection{Dataset}
\subsubsection{Comparison to Existing Data}
\begin{figure*}[!h]
    \centering
    \vspace{-5mm}
    \includegraphics[width=.9\textwidth]{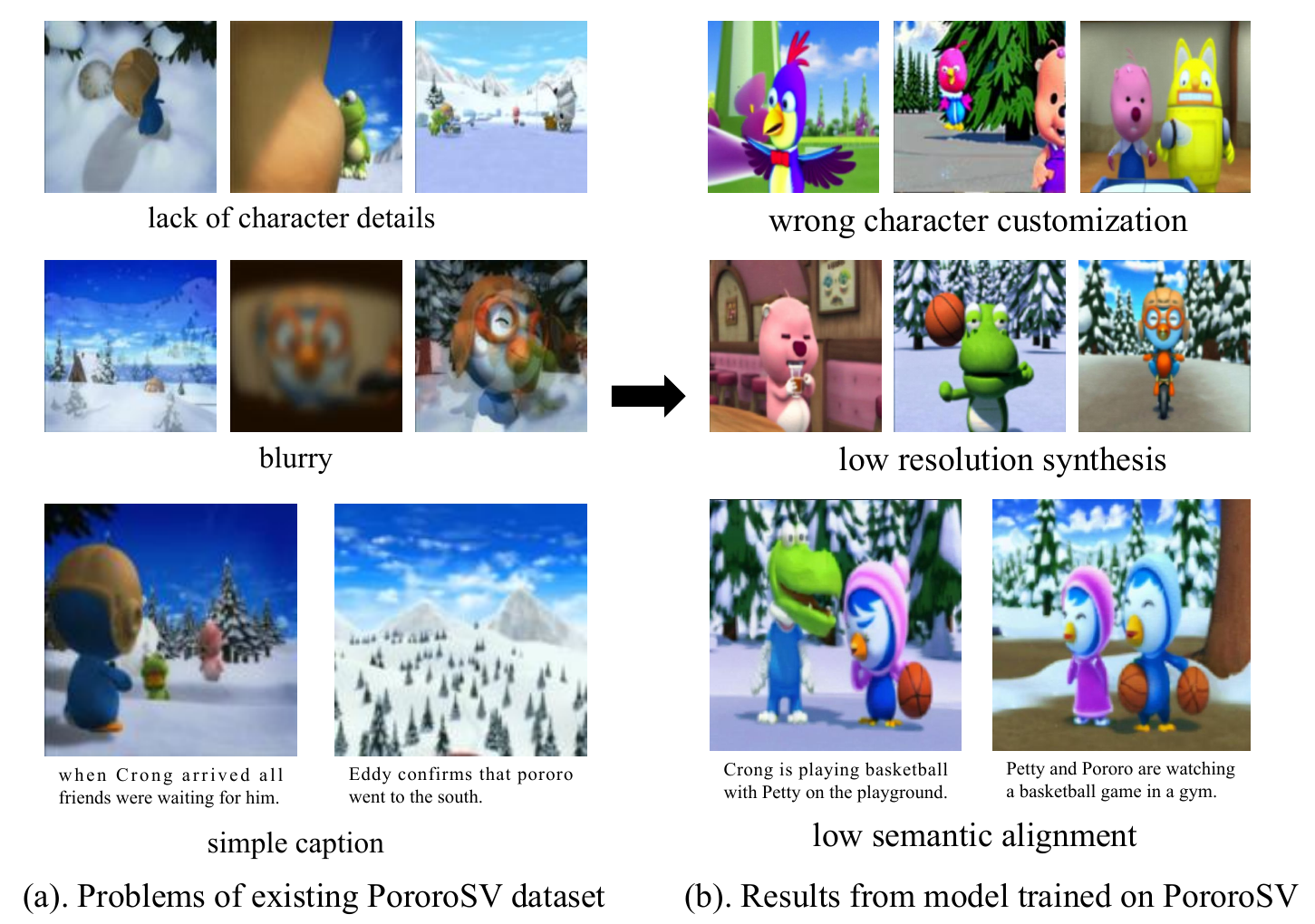}
    \caption{The current datasets PororoSV used for story visualization faces challenges like low resolution, blurry training samples and caption annotations. These factors combined impede the model from achieving high-quality character customization.
    }
    \label{fig:existing_dataset}
\vspace{-3mm}
\end{figure*}

PororoSV ~\cite{li2019storygan} is a commonly used dataset for story visualization, comprising a total of 73,665 frames featuring 9 characters. However, through our experiments, we discovered that this dataset is not ideal for character customization due to three main reasons, as shown in Fig.~\ref{fig:existing_dataset}(a). Firstly, some training images from PororoSV lack character details due to perspectives like overlooking or distant viewpoints. Secondly, the low resolution \emph{i.e.}, 128x128, and frame shifts contribute to blurriness in each sample. Finally, as the original PororoSV dataset is primarily intended for video question answering, its image captions may be simplistic and insufficient to fully describe the depicted content.


As a result, when the model is trained on this dataset, it faces numerous challenges and struggles to achieve high-quality character customization. The scarcity of character details within the dataset can lead to difficulties in achieving accurate character customization, potentially resulting in the generation of new characters that bear only a rudimentary resemblance to the ground truth character. Moreover, the synthesized images may be limited to low-resolution generation with insufficient semantic alignment, as shown in Fig.~\ref{fig:existing_dataset}(b).
\begin{figure*}[!h]
\centering
\vspace{-5mm}
\subfigure[Total Character in our experiments]{
\includegraphics[width=.99\linewidth]{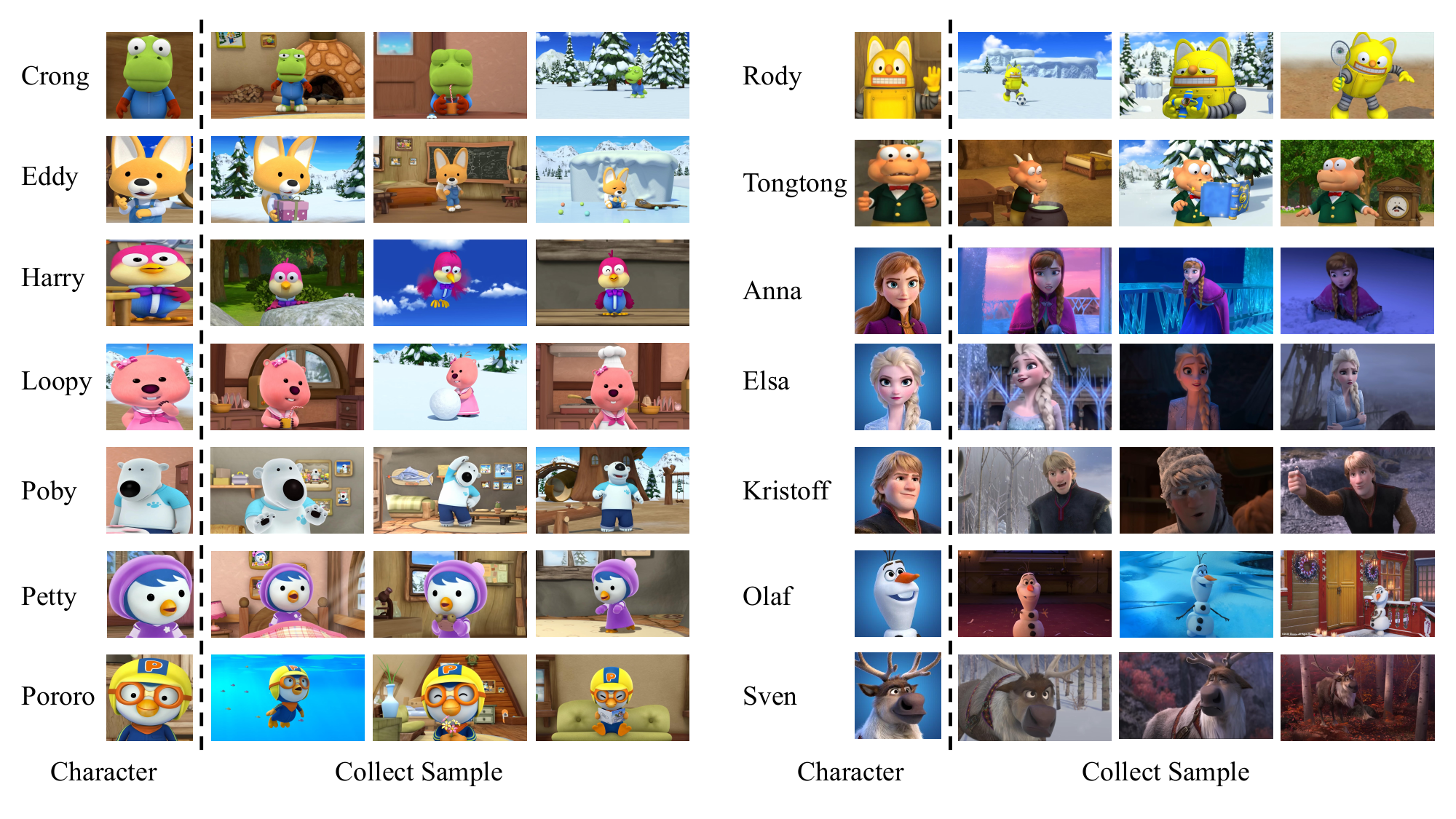} 
}
\subfigure[Examples for multi-character collection]{
\includegraphics[width=.99\linewidth]{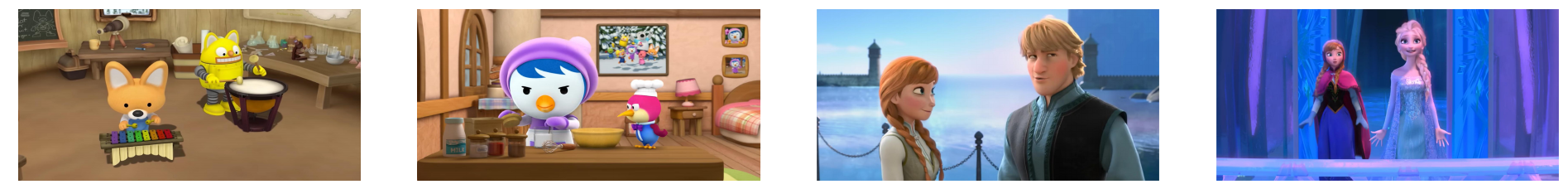} 
}
\vspace{-3mm}
\caption{Examples of our collected characters and samples focus on two cartoon series, namely \emph{i.e.}, Pororo and Frozen. These samples include detailed depictions of individual characters and scenarios that showcase interactions among multiple characters.}
\label{fig:ours_dataset}
\end{figure*}

To this end, we construct a high-quality dataset containing various events tailored for character customization and story visualization. We collect story frames from open-source websites, focusing not only on Pororo but also on another renowned cartoon series, ``Frozen." The character lineup for each cartoon is depicted in Fig. \ref{fig:ours_dataset}(a), with a total of 9 characters for Pororo and 5 for Frozen. Due to resource constraints, we curate around 10 images depict only single character for each character and over 30 frames for scenarios involving multiple characters. All samples are collected from open-source websites\footnote{We collect the training samples from \url{https://www.youtube.com/@pororoenglish} and  \url{https://www.youtube.com/@DisneyUK}}. Moreover, considering the resolutions of the collected image samples, \emph{i.e.}, 1080 $\times$ 1920, we adapt the diffusion model to synthesize images at a resolution of 512 $\times$ 896 to ensure the consistent generation quality.

We also introduce a newly validation dataset designed for the two customized cartoon characters. For single-character customization, we prompt ChatGPT to craft 5 stories for each character, with each story comprising 9 unique scenes. In terms of multi-character story visualization task, we create a total of 10 stories for each cartoon, encompassing 9 diverse scenes involving multiple characters. We generate 5 images per scene for every story across all compared methods, evaluating the metrics based on the comprehensive set of generated results to ensure a fair comparison.

\subsubsection{Character Vocabulary Construction}
To construct Character-Graph, we first build a character vocabulary. This vocabulary comprises the  frontal image of each character as the key and the corresponding character maps representing the appearance of that particular character as the associated values.

\begin{figure*}[!h]
    \centering
    \includegraphics[width=1.\textwidth]{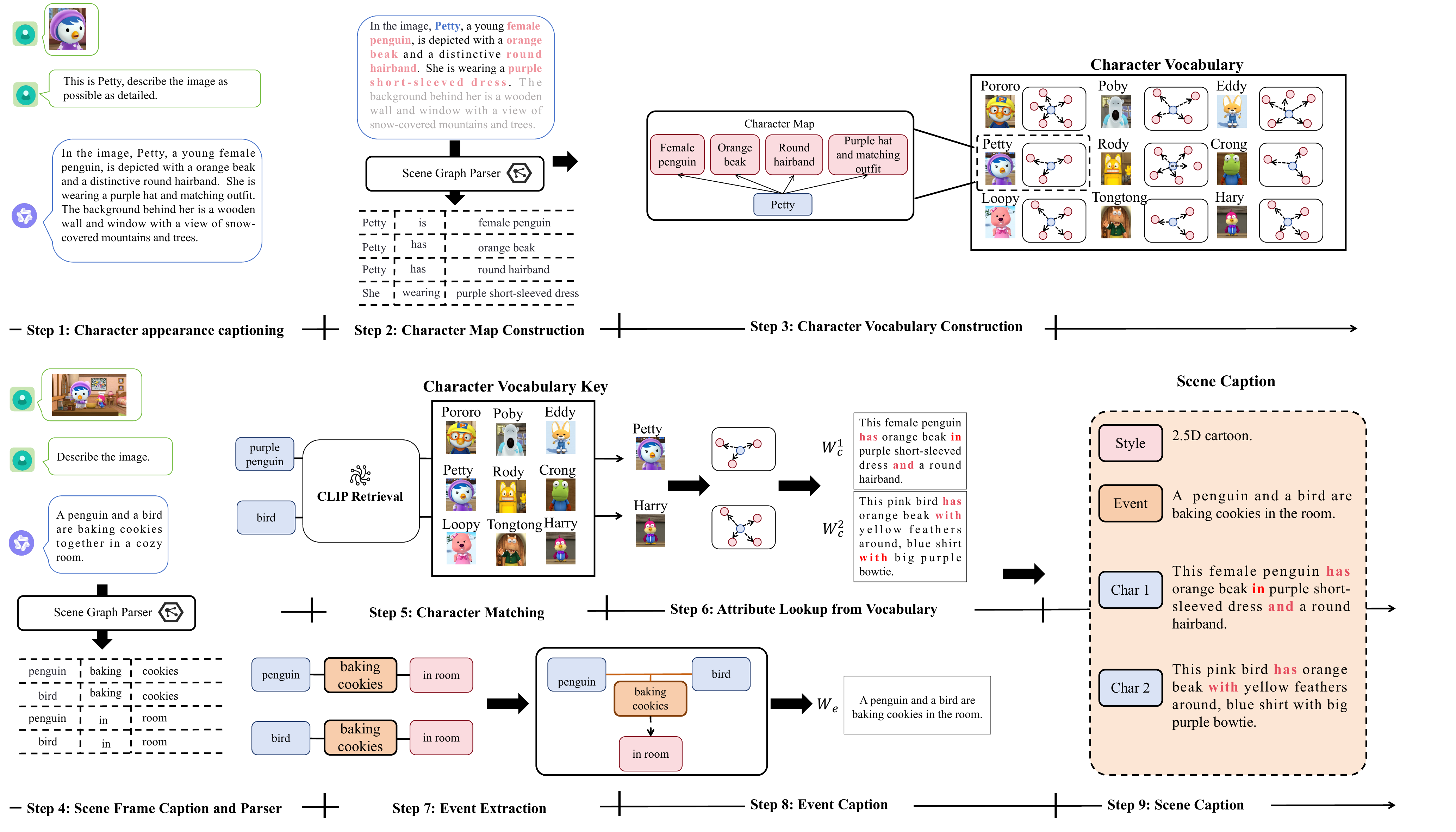}
    \caption{The detailed Character-Graph Construction and Scene Caption.
    }
    \label{fig:C-CG}
\vspace{-3mm}
\end{figure*}

To create the character map for each character, we initially input the frontal image of the character and query a Vision Language Model (VLM) \emph{i.e.}, Qwen-VL \cite{ali2023qwenvl}, to provide a detailed description of the character's appearance (\textbf{Step 1} in Fig.\ref{fig:C-CG}). However, the VLM's response may include descriptions of unrelated regions in the image, such as the background. To address this, we employ a Scene Parser to extract the entities, attributes, and relationships depicted in the response. Subsequently, we choose the distinctive appearance features for each character to construct a character map $Map<O_i, A_i^k>$ (\textbf{Step 2} in Fig.\ref{fig:C-CG}). For example, ``Petty'' is represented by $O_i$ and each attribute description like ``female penguin''is denoted as $A_i^k$. Then for every character 
 $O_i$ in the character set, we generate $Map<O_i, A_i^k>$ and establish a Character Vocabulary (\textbf{Step 3} in Fig.\ref{fig:C-CG}).

\subsubsection{Scene Caption via Character Graph}
As mentioned in the main paper, to generate a semantically rich caption for each story frame $\mathcal{F}$, we first represent it using the Character Graph, which is then converted into the text prompt $T_{g} = [W_{s}, W_{e}, \sum_j^N W_{c}^{j} ]$. Specifically, we query the VLM to describe the content of $\mathcal{F}$. Then, the response is parsed by a Scene Parser to extract the subject (character), relationships (event), and attributes, as depicted in \textbf{Step 4} in Fig. \ref{fig:C-CG}. Following, we will delve into the detailed construction of the character description $W_{c}^j$, event description $W_e$ and style description $W_s$ through C-CG.

\subsubsection{Character Description} To generated character description $W_c^{j}$ for each character $char_j$, we obtained the coarse class label of the character, \emph{e.g.}, ``purple penguin'' and ``bird'' from the scene graphs generated by the Scene Parser as demonstrated in \textbf{Step 4} Fig.~\ref{fig:C-CG}. We then use a CLIP model to compute the similarity between this coarse class label and the keys in the Character Vocabulary, \emph{i.e.}, frontal images for the characters. The character with the highest similarity is recognized as the matched character. This retrieval and matching process can be formulated as:
\begin{small}
\begin{equation}
\begin{aligned}
O_{char}^j  =  \mathop{\arg\max}\limits_{O_i} Sim(char_j, O_i).
\end{aligned}
\end{equation}
\end{small}%
Here, $char_j$ represents the coarse class label, and $O_i$ is a frontal image in the Character Vocabulary and $Sim$ is the CLIP feature similarity calculation. This process is detailed in \textbf{Step 5} of Fig.~\ref{fig:C-CG}. 

Then we refer to the constructed character vocabulary to retrieve the associated attributes, formulated as:
\begin{small}
\begin{equation}
\begin{aligned}
A_{char}^{j}=  O_{char}^j \otimes  \sum_{k} Map<O_i, A_i^k>,
\end{aligned}
\end{equation}
\end{small}%
where $O_{char}^j$ denotes the matched character, and $Map<O_i, A_i^k>$ represents each related attribute related to $O_{char}^j$. The operation $\otimes$ is executed through a dictionary lookup. Finally, we craft the character description for $char_j$ as:
\begin{small}
\begin{equation}
W_c^{j} = O_{char}^j \oplus A_{char}^j.
\end{equation}
\end{small}%
where we combine the character name with its unique attributes using connecting words like "and" "with," or "in" (\textbf{Step 6} in Fig. \ref{fig:C-CG}).

\subsubsection{Event Description} Through the Scene Parser, we are also able to extract the subjects and their relationships from the scene graph.  Then we merge the graphs by identifying identical relationships, \emph{e.g.}, ``baking cookies'' in the provided example in Fig.\ref{fig:C-CG}, and create event description $W_e$ by:
\begin{small}
\begin{equation}
W_e = \sum_{j, j^{\prime}} O_{char}^j \oplus R_{{j, j^{\prime}}} \oplus O_{char}^{j^{\prime}}
\end{equation}
\end{small}%
Here, $R_{j, j^{\prime}}$ represents the same event performed by two characters $O_j$ and $O_{j^{\prime}}$. Similarly, we merge the objects and their relationships using connecting words like "and" or "together" (\textbf{Step 8}). It's important to note that when there is only one character in the image, \emph{i.e.}, $O_j = O_{j^{\prime}}$, we exclude $O_{j^{\prime}}$ from $W_c$ and $W_e$ in that scenario.

\subsubsection{Style Description} To enrich the comprehensiveness of the Character-Graph, we introduce a descriptive sentence to delineate the style of all scenes within the story's world, denoted as $W_s$. For cartoon \emph{Pororo the Little Penguin}, $W_s$ is defined as ``2.5D Cartoon'', whereas for cartoon \emph{Frozen}, $W_s$ is ``Disney movie style.''

Through the aforementioned process, each frame $\mathcal{F}$ can be represented via Character-Graph $G(O, E) = \{ \sum_{i, k} Map<O_i, A_i^k>, \sum_{i, j} R(O_i, O_j)\}$ and captioned as $T_g$. This enhancement enables the customization of all characters through a unified diffusion model.

\subsection{Comparison Methods}
\subsubsection{StoryGEN} 
StoryGEN ~\cite{liu2024intelligent} achieves open-ended visual storytelling by conditioning diffusion models on both the current text prompt and preceding image-text pairs. To extract contextual features from previous frames, they introduce a visual-language context module to extract contextual features from previous frames. The contextual semantics are encoded by leveraging a pre-trained stable diffusion model to denoise preceding frames that have been corrupted by noise, following guidance from the corresponding captions. Image features after each self-attention layer in the UNet blocks are used as the conditioning visual context feature. To fuse information from the current text prompt with the conditioning visual context feature, they enhance the transformer decoder in SDM with an additional image cross-attention mechanism. The total training strategy comprises a two-stage approach, including single-frame pre-training and multi-frame fine-tuning. This strategy aims to ensure proficiency in single-frame generation and contextual features extraction for multi-frame generation. They also curate a dataset named StorySalon, comprising diverse videos and E-books, and use TextBind ~\cite{li2023textbind} to generate captions for each image. 

As StoryGEN primarily focuses on achieving open-ended story visualization, it places less emphasis on precise character identity customization. Moreover, the captions in StorySalon is simple and lack of detailed story semantics. Consequently, StoryGEN faces challenges in generating new stories based on provided characters with high identity preservation and accurate semantic alignment.

In our experiment comparison, we use the official implement code\footnote{\url{https://github.com/haoningwu3639/StoryGen}} and their released checkpoint for both \textbf{single-character} and \textbf{multi-character} story visualization. To achieve pre-defined character-based story visualization, we provide the collected character images of the specific character that appears in the current frame as additional preceding images to the model in each plot generation.



\subsubsection{IP-Adapter} 
IP-Adapter ~\cite{ye2023ip} introduces an efficient adapter for guiding text-to-image generation with image prompts. They leverage a pre-trained CLIP image encoder to extract features from the provided image prompt, aligning these features with text features from a pre-trained diffusion model through a trainable projection network. Subsequently, they introduce a decoupled cross-attention mechanism to separate the cross-attention layers for text and image features, respectively. Trained on a large dataset, IP-Adapter enables image-guided text-to-image generation.  However, as the extracted image features are primarily focus on semantic information, IP-Adapter lack of detailed appearance awareness necessary for faithful character identity reconstruction. To address this problem, a new version of IP-Adapter, termed IP-Adapter-plus is proposed. IP-Adapter-plus use Q-Former~\cite{li2023blip} to represent the input image with more detailed information using multiple image tokens. While IP-Adapter-plus enhances subject identity preservation, it compromises text-semantics alignment. 

We employ the official implementation code\footnote{\url{https://github.com/tencent-ailab/IP-Adapter}} and two released models \emph{i.e.}, IP-Adapter-Based model and IP-Adapter-plus model, for \textbf{single-character} story visualization. Similar to StoryGEN, the frontal image of the pre-defined character are served as the referenced image for character-guided text-to-image generation. The image guidance scale and text guidance scale are set to default values.

\subsubsection{Dreambooth}
Dreambooth ~\cite{ruiz2023dreambooth} resort to fine-tuning the entire diffusion model on a small image set containing personalized subjects for image customization. They use rare tokens as identifiers to bind the specific subject,wherein a text prompt consists of a unique identifier followed by the character's class name is used as the text caption to train the model for subject reconstruction. However, the reliance on a sole token for subject customization in Dreambooth leads to capturing irrelevant information from the training samples, impeding  effective character customization. As a result, Dreambooth often struggles to maintaining a delicate balance between preserving the subject's identity faithfully and ensuring semantic coherence. We use the code of diffusers\footnote{\url{https://github.com/huggingface/diffusers}} to train Dreambooth individually for each character for \textbf{single-character} story visualization.To enhance both the quality of output and the speed of convergence, we also fine-tune the text-encoder together with the Unet.

\subsubsection{LoRA} 
 By enhancing pre-trained models with pairs of rank-decomposition matrices added to existing weights and training only these new weights, LoRA ~\cite{hu2021lora} avoids catastrophic forgetting, achieving superior text semantic alignment with better portability compared to Dreambooth. However, due to a smaller learnable parameter space, LoRA exhibits limited identity customization using the same special tokens. Overall, both LORA and Dreambooth face similar challenges in maintaining well-matched story visualization alongside high-fidelity identity preservation. We leveraged the diffusers library to implement LORA for each character in \textbf{single-character} story visualization.

\subsubsection{Mix-of-Show} 
Mix-of-Show ~\cite{gu2024mix} adopts an embedding-decomposed LoRA for single-client tuning and gradient fusion for the central node to preserve the in-domain essence of individual concepts. To enable multi-concept customization, Mix-of-Show further introduces regionally controllable sampling within the fused LoRA , expanding spatially controllable sampling to address attribute binding and missing object issues in multi-concept sampling. Specifically, region-based sampling necessitates users to input a sketch pose image along with a layout to specify the region where a character will be generated. However, due to these strict constraints, Mix-of-Show encounters layout conflicts when the text prompts pose a greater challenge for interactive generation. In line with Mix-of-Show, we utilize Anything-v4 as the pre-trained model and train the ED-LORA using their official code\footnote{\url{https://github.com/TencentARC/Mix-of-Show}} in terms of \textbf{multi-character} visualization. All parameters align with those provided in the codes.

\subsubsection{LoRA-Composer} 
LoRA-Composer ~\cite{yang2024lora} tackles the issue of concept vanishing in LoRAs merging by concept injection constraints. Specifically, they incorporate features from each LoRA corresponding to individual characters using a regionally sampling technique akin to Mix-of-Show, based solely on an input layout. To preserve the intricate details of each customized concept, they introduce concept enhancement constraints through an expanded cross-attention mechanism. However, this method faces challenges in generating natural interactions due to the input layout constrains. Moreover, as LoRA-Composer mandates each LoRA to represent a unique concept rather than merging them into a unified model, it necessitates a larger number of parameters to construct a multi-character story. We leverage the ED-LoRA trained by Mix-of-Show and use the official sampling code\footnote{\url{https://github.com/Young98CN/LoRA_Composer}} to achieve \textbf{multi-character} visual storytelling.

\subsection{Evaluation Metric}
Following previous work ~\cite{liu2024intelligent,ruiz2023dreambooth,gong2023talecrafter,gu2024mix,yang2024lora}, we evaluate the quality of the generated image series based on two main criteria: Character-Identity Preservation and Text-Semantic Alignment. For single-character story visualization, we employ CLIP-I and DINO-I to assess character-identity preservation. When it comes to multi-character generation, we consider Frame-Accuracy and Character F1 scores. For text-semantic alignment, we use CLIP-T. We generate 5 images per scene for every story across all compared methods, evaluating the metrics based on the comprehensive set of generated results to ensure a fair comparison. Below we we will delve into each metric and elucidate their respective implementations.
\subsubsection{CLIP-I \& DINO-I}
CLIP-I measures the average pairwise cosine similarity between the ViT-B/32 CLIP embeddings of the generated images and the GT images. For adapter-based methods, we compute the mean cosine similarity between the generated images and their corresponding reference images. For customization-based methods, we first calculate the mean cosine similarity of the generated images with the corresponding customized training samples. Then an average similarity is obtained by computing the mean similarity for each sample. A higher CLIP-I score indicates a more pronounced semantic alignment between the synthesized image and the reference images.
\begin{table*}[!h]
\centering
\footnotesize{
\renewcommand{\arraystretch}{1.1}
\setlength{\tabcolsep}{1.4mm}{
\begin{tabular}{c|ccc|ccc|ccc|ccc}
\hline
\multirow{2}{*}{\textbf{Setting}} & \multicolumn{6}{c|}{\textbf{Pororo}} & \multicolumn{6}{c}{\textbf{Frozen}} \\
\cline{2-13}
& \multicolumn{3}{c|}{\textbf{Single Character}} & \multicolumn{3}{c|}{\textbf{Multi Character}} & \multicolumn{3}{c|}{\textbf{Single Character}} & \multicolumn{3}{c}{\textbf{Multi Character}}\\

& \textbf{DINO-I} & \textbf{CLIP-I} & \textbf{CLIP-T}& \textbf{CLIP-T} & \textbf{F-Acc} &\textbf{C-F1} &\textbf{DINO-I} & \textbf{CLIP-I} & \textbf{CLIP-T}& \textbf{CLIP-T} & \textbf{F-Acc} &\textbf{C-F1}\\
\hline

w/o C-CG \& KE-SG &39.57 & 68.11 &28.74 & 30.98 & 22.22& 48.26 &49.74 & 76.97 &27.16 & 28.63 & 21.54& 32.14\\
\hline
\textcolor{gray}{dreambooth} &\textcolor{gray}{61.85} & \textcolor{gray}{78.86} &\textcolor{gray}{26.74} & - & - &  -  &\textcolor{gray}{55.01} & \textcolor{gray}{81.07} & \textcolor{gray}{27.12} & - & - &  - \\
w/o KE-SG &60.98 & 81.28 &32.17 & 33.41 & 35.73&  52.69 &59.98 & 81.80 &34.95 & 32.76 & 31.65& 38.21\\
\hline
StoryWeaver &\textbf{64.96} & \textbf{82.65 }&\textbf{33.26} & \textbf{34.30} & \textbf{40.45}& \textbf{59.72} &\textbf{62.17} & \textbf{85.24} &\textbf{36.74} & \textbf{34.94} & \textbf{34.51}& \textbf{44.53}\\
\hline
\end{tabular}
\vspace{-2mm}
\caption{Ablation study of on Pororo. \emph{\textbf{C-CG}} refers to customization via Character-Graph and \emph{\textbf{KE-SG}} refers to the knowledge-enhanced spatial guidance. 
}
\label{tab:full_ablation}
}

}
\end{table*}

\begin{table}[!t]
\centering
\footnotesize{
\renewcommand{\arraystretch}{1.1}
\setlength{\tabcolsep}{2.mm}{
\begin{tabular}{c|ccc|ccc}
\hline
\multirow{2}{*}{\textbf{Methods}} & \multicolumn{3}{c|}{\textbf{Pororo}} & \multicolumn{3}{c}{\textbf{Frozen}} \\
\cline{2-7}
& \multicolumn{6}{c}{\textbf{Single Character Generation}} \\
& \textbf{T-A} & \textbf{I-A} & \textbf{V-Q}& \textbf{T-A} & \textbf{I-A} & \textbf{V-Q}\\
\hline
StoryGEN & 1.34 & 1.42 & 1.00 & 1.30  & 1.20 & 1.22  \\
IP-Adapter (base) & 3.12 &2.12  &2.54  & 2.90  & 2.08 & 2.68  \\
IP-Adapter (plus) & 2.22 &3.92  & 3.57 & 2.22  & 3.54 &3.0   \\
\hline
LoRA & 2.84 & 2.16 & 2.46 & 2.58  &1.72  & 2.66  \\
Dreambooth & 2.48 & 3.26 & 2.54 & 2.46  & 3.12 &2.58   \\
\hline
Ours & \textbf{4.12} & \textbf{4.10} &\textbf{4.36}  & \textbf{3.98}  &\textbf{4.16}  &\textbf{4.34}   \\
\hline
\hline
\multirow{2}{*}{\textbf{Methods}}& \multicolumn{6}{c}{\textbf{Multi Character Generation}} \\
& \textbf{T-A} & \textbf{I-A} & \textbf{V-Q}& \textbf{T-A} & \textbf{I-A} & \textbf{V-Q}\\

\hline
StoryGEN & 1.24 &1.62 & 1.50 & 1.48  & 1.54 &1.36   \\
\hline
Mix-of-Show & 2.80 & 3.46 &3.54  &2.68   & 3.48 & 3.34  \\
LoRA-Composer & 3.22 &3.52  & 4.08 &3.28   & 3.48 &3.98   \\
\hline
Ours & \textbf{4.22} &\textbf{4.42}  & \textbf{4.26}  &\textbf{4.28}   &\textbf{4.20}  &  \textbf{4.38} \\
\hline
\end{tabular}
}
\vspace{-2mm}
\caption{Complete results of user Study. A higher score indicates better performance in terms of the corresponding metric. Our StoryWeaver again achieve the best performance in terms of three aspects.
}
\label{tab:full_user_study}
}
\vspace{-3mm}
\end{table}
However, as illustrated in ~\cite{ruiz2023dreambooth}, CLIP-I lacks the ability to differentiate between distinct characters with highly similar features. Following \citep{ruiz2023dreambooth}, we introduce DINO-I as an additional metric for evaluating identity preservation. DINO is trained using a self-supervised objective that emphasizes distinguishing unique features of a subject, making it more suitable for identity preservation assessment as it focuses more on the object itself. DINO-I represents the average pairwise cosine similarity between the ViT-G14 DINO embeddings of the generated images and the real images corresponding to the customized characters. Similar to CLIP-I, we calculate the mean cosine similarity between the generated images and the corresponding reference image in adapter-based methods, and the average similarity of the mean DINO-I score of each generated sample with its customized image sets. A higher DINO-I score indicates an improved alignment of identity between the synthesized image and the reference images.

\subsubsection{CLIP-T}
CLIP-T measures the similarity between the features of the generated image and the input text in the CLIP feature space. We compute the average CLIP-T by ViT-B/32 CLIP for each story scene and its corresponding text prompt to assess the semantic alignment of each generated visual narrative. A higher CLIP-T score indicates superior semantic alignment.

\subsubsection{Character-F1}
The Character F1 score evaluates the proportion of characters in the images that precisely correspond to the characters in the story text inputs. The Character F1 score for each generated sample is computed as follows:
\begin{equation}
    \text{Character-F1} = \frac{Match_i}{N_t},
\end{equation}
where $N_t$ represents the total number of characters mentioned in the input text prompt, and $Match_i$ is the count of characters that are exact matches in that particular scene.

In practice, we use DINO as a character classifier. We extract the image features from all images within the ground truth (GT) character set. Subsequently, for each generated sample, we extract its image features and compute the cosine similarity with the GT character image features. The classification result with the highest similarity score is considered valid only if it surpasses 0.5. The Character-F1 is then computed based on these valid classifications. Finally, we calculate the average Character-F1 across all image sets to derive the mean Character-F1. A higher Character-F1 score indicates improved identity preservation in multi-character generation scenarios.

\subsubsection{Frame-Accuracy}
While the Character F1 score assesses the percentage of characters depicted in a narrative, Frame Accuracy measures the proportion of instances where all characters are correctly included in a story. Similar to Character F1, we utilize DINO as a character classifier to ascertain the presence of characters in the image frame. The Frame Accuracy is determined as follows:
\begin{equation}
    \text{Frame-Acc} = \frac{Match_f}{N_f},
\end{equation}
Here, $N_f$ represents the total number of frames in the story, and $Match_f$ signifies the number of correct frames that contain all the characters required for that scene. A higher Frame-Acc score signifies better quality in generating multi-character narratives.

\subsubsection{Model Parameters}
We also incorporate the model parameters as an additional metric in the comparison. Specifically, we compute the total parameters per dataset involved in a single forward generation, including the VAE encoder, text encoder, UNet, and the specialized modules for each technique, \emph{e.g.}, image encoder and image projection network for IP-Adapter. For LORA-based approaches, we solely consider the additional parameters for a new character besides the base modules (VAE encoder, text encoder and UNet). As illustrated in the main paper, through the Character-Conditioned Graph (C-CG) and the Knowledge-Enhanced Scene Graph (KE-SG), StoryWeaver demonstrates its ability for high-quality story visualization in various scenarios within \textbf{a unified model}, showcasing exceptional efficiency in model storage capacity.
\subsubsection{User Study}
\begin{figure*}[t]
\centering
\vspace{-6mm}
\subfigure[Examples for Single-Character Generation]{
\includegraphics[width=.89\linewidth]{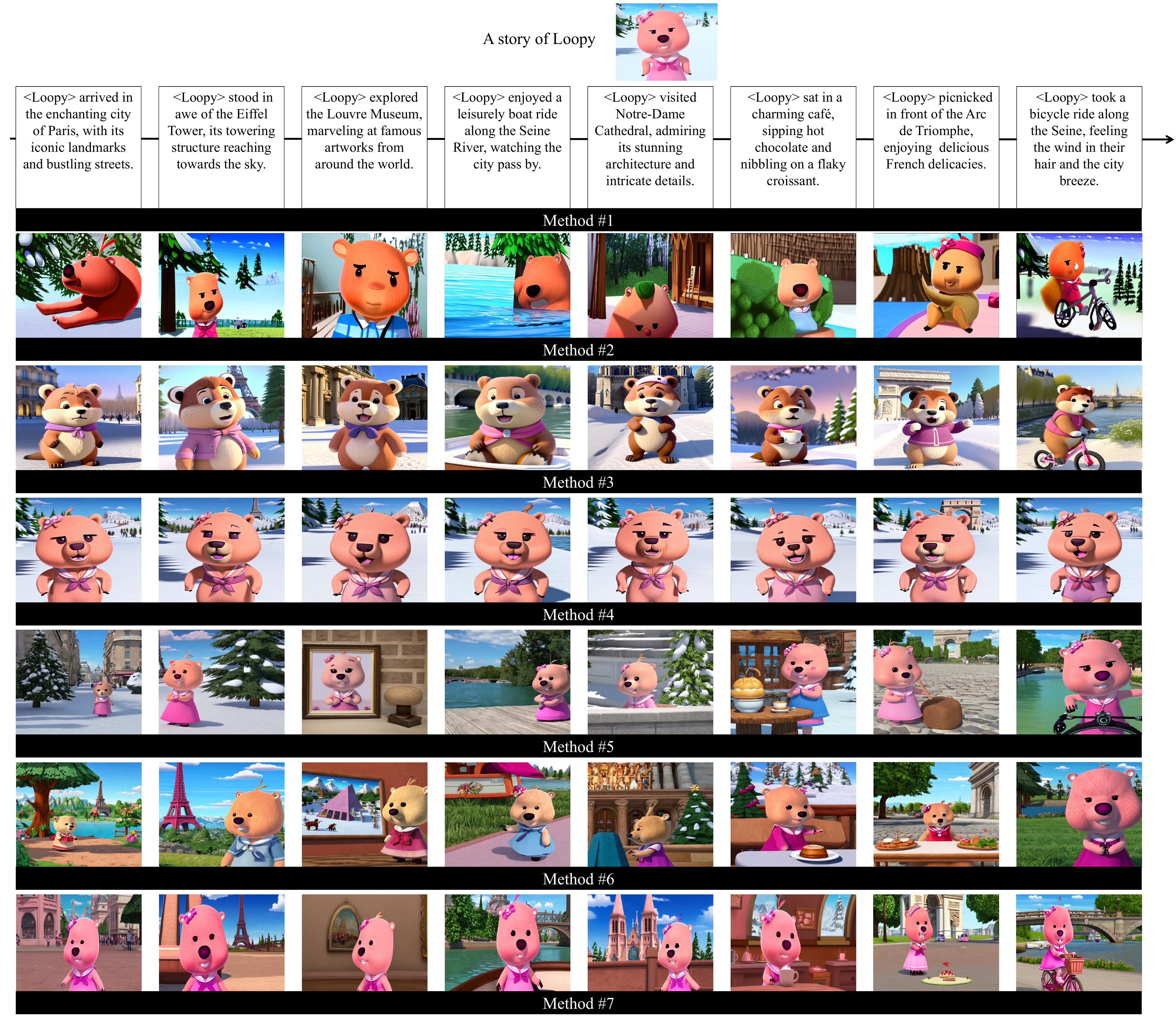} 
}
\subfigure[Examples for Multi-Character Generation]{
\includegraphics[width=.89\linewidth]{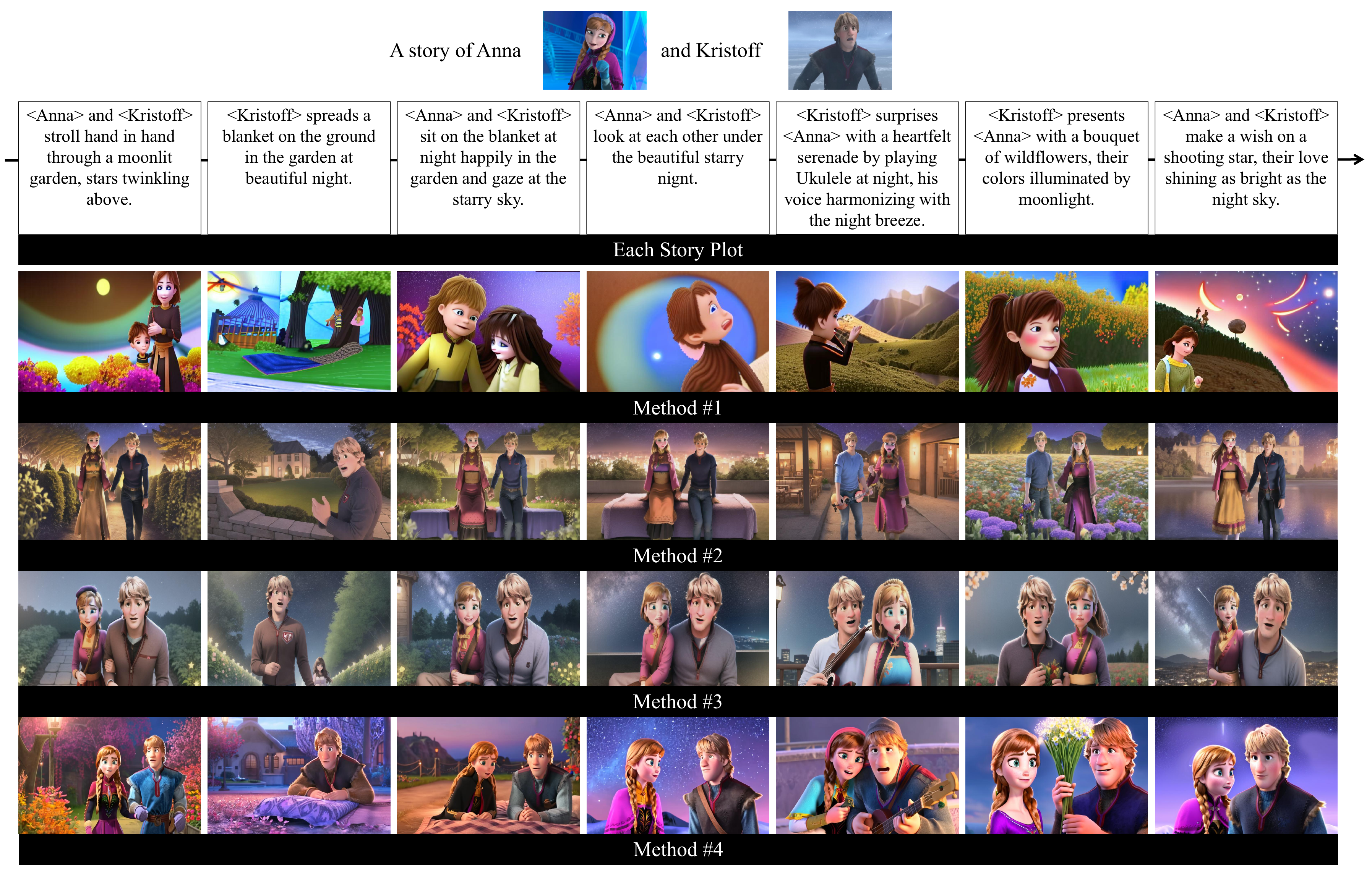} 
}
\vspace{-3mm}
\caption{Examples of our user study. We present the results for both single- and multi-character story visualization anonymously to users and ask them to rank the methods based on the introduced three criteria.}
\label{fig:user_study}
\end{figure*}

We conduct a user study with other exiting methods for both single-character and multi-character generation. Specifically, we invite 50 participants to rank all methods based on three criteria: (1) \emph{Text-Alignment} (\textbf{T-A}), ranking the methods based on the alignment between generated story scene and the input text narrative; (2) \emph{Image-alignment} (\textbf{I-A}), ranking the methods based on the identity preservation of the generated image with the customized character frontal image as the GT image; and (3) \emph{Overall Visual
Storytelling Quality} (\textbf{V-Q}), ranking the methods based on the perceptual quality. We asked the participants to assign scores from 5 to 1 for each ranking, where a higher score indicate a better performance. We randomly select five single-character visual stories and five multi-character visual stories for each dataset. As shown in Fig.~\ref{fig:user_study}, the results are presented anonymously to the participants.

\subsection{Additional Result}
\subsubsection{Ablation Study}
We present the complete results of the ablation study conducted on both datasets in Tab.\ref{tab:full_ablation}. The introduction of Customization via Character-Graph significantly enhances character identity preservation, as demonstrated by the improvements in DINO-I and CLIP-I for both datasets. Furthermore, the comprehensive scene captions generated by the Character-Graph enable our StoryWeaver to achieve superior performance compared with the single-concept customization method, \emph{i.e.}, Dreambooth. Through our proposed Knowledge-Enhanced Spatial Guidance, StoryWeaver continues to enhance the quality of multi-character generation, exhibiting higher scores in Frame Accuracy and Character F1. This further validates the efficacy of KE-SG in mitigating identity blending issues. Overall, both proposed designs enhance the model and facilitate unified character customization with improved performance.

\begin{table}[!t]
\centering
\footnotesize{
\renewcommand{\arraystretch}{1.5}
\setlength{\tabcolsep}{2.5mm}{
\begin{tabular}{c|cc|cc}
\hline
\multirow{2}{*}{\textbf{Metric}} & \multicolumn{2}{c|}{\textbf{p-values}} & \multicolumn{2}{c}{\textbf{ICC}} \\
\cline{2-5}
& Sin-char & Multi-char & Sin-char & Multi-char \\
\hline
\textbf{T-A} & 1.71e-5 & 3.00e-8&0.87 &0.96\\
\hline
\textbf{I-A} & 4.96e-7 & 5.80e-5&0.91 &0.86\\
\hline
\textbf{V-Q} & 6.06e-6 & 4.81e-6&0.88 &0.94\\
\hline
\end{tabular}
}
\vspace{-2mm}
\caption{Results for statistic significance and inter-class correlation coefficient of each metric.
}
\label{tab:human_evaluation}
}
\vspace{-3mm}
\end{table}
\subsubsection{User Study}
In Table \ref{tab:full_user_study}, we present the complete results for both single- and multi-character generation on the two datasets in Table \ref{tab:full_user_study}. Our StoryWeaver exhibits remarkable performance in text semantic alignment, identity preservation, and visual quality, surpassing existing methods across both datasets. This result further underscores the effectiveness of our proposed techniques.

As shown in Table \ref{tab:full_user_study}, the rating scores of different methods vary significantly, highlighting a clear user preference under fair comparison. We further present the statistical significance and inter-annotator agreement, of which the results are shown in Table ~\ref{tab:human_evaluation}. 

In terms of statistical significance, results were tested against the null hypothesis that participants had no preference among methods with equal scores for all methods for each metric. The p-values are calculated by Analysis of Variance (ANOVA), and we choose p< 0.05 as a significant deviation from the null hypothesis. $\chi^2$ tests were also conducted under the same null hypothesis. All p-values were well below 0.05, indicating a strong preference for the visual results of StoryWeaver. As for Inter-annotator Agreement, we evaluate by Inter-class Correlation Coefficient (ICC). The results also confirm a consistent preference for StoryWeaver across different metrics.

\subsubsection{Additional Comparison}
\begin{figure*}[h]
\centering
\vspace{-5mm}
\subfigure[]{
\includegraphics[width=1.\linewidth]{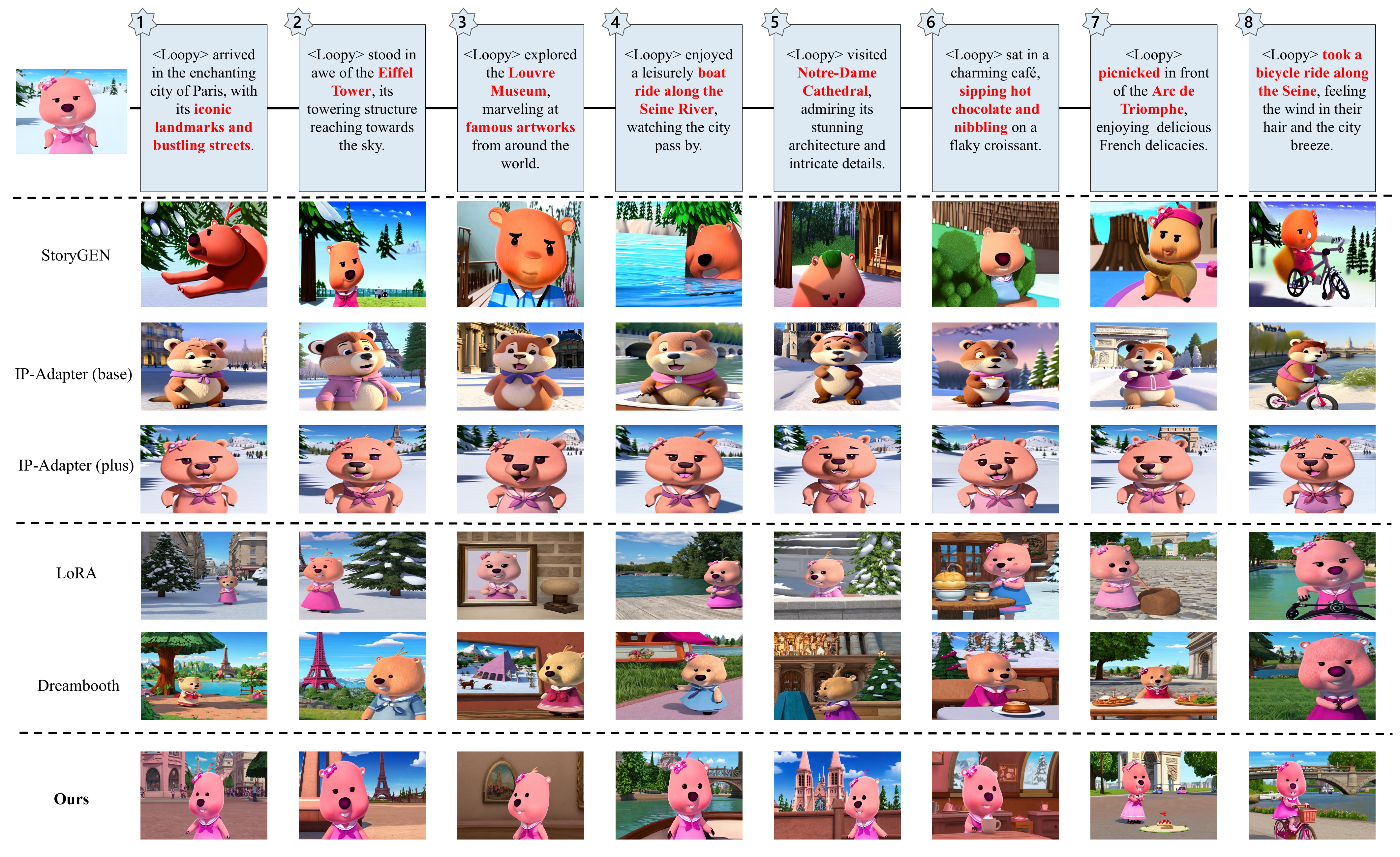} 
}
\subfigure[]{
\includegraphics[width=1.\linewidth]{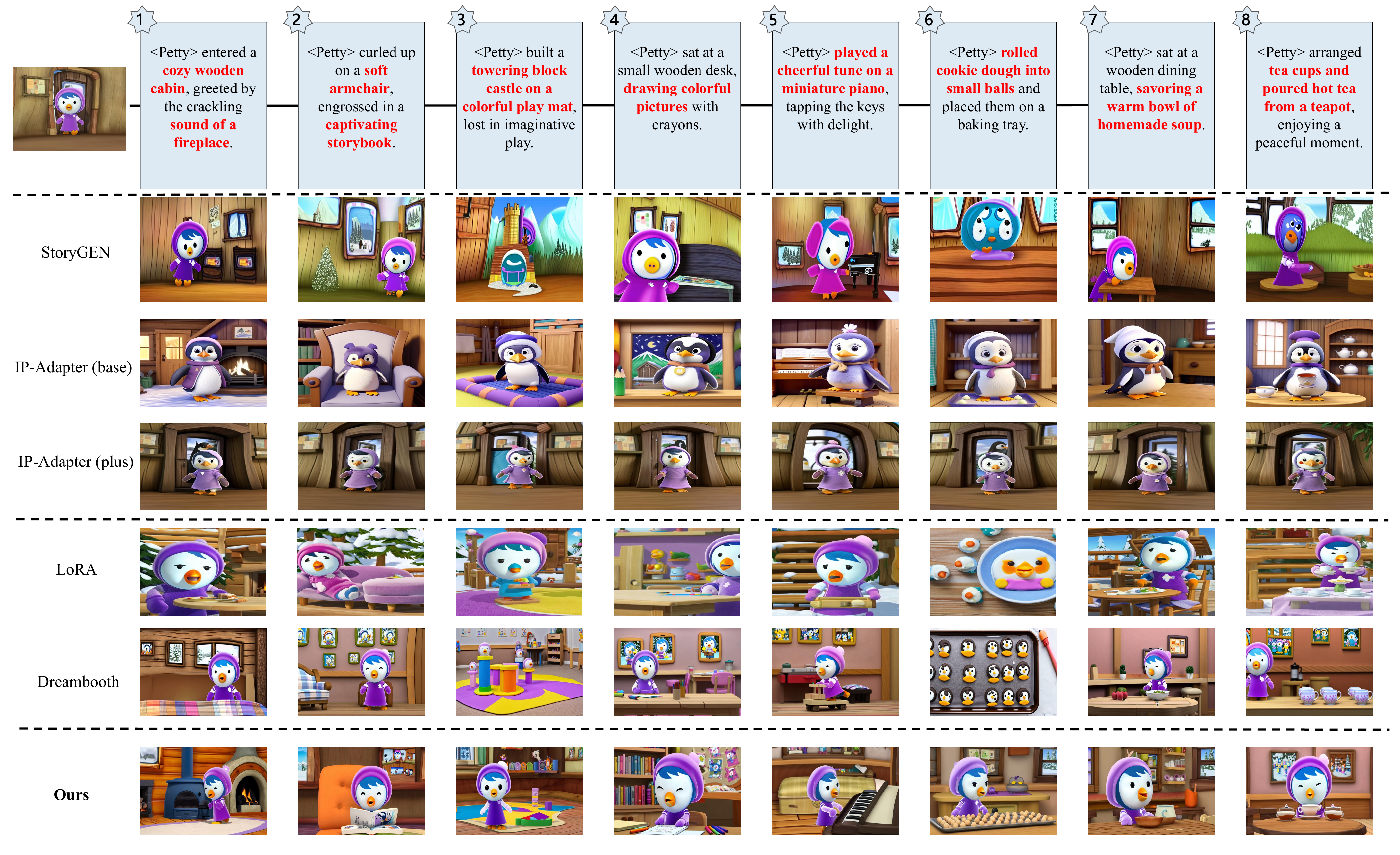} 
}
\vspace{-3mm}
\caption{Examples for single-character story visualization on Pororo Dataset.}
\label{fig:single-pororo}
\end{figure*}

\begin{figure*}[h]
\centering
\vspace{-5mm}
\subfigure[]{
\includegraphics[width=1.\linewidth]{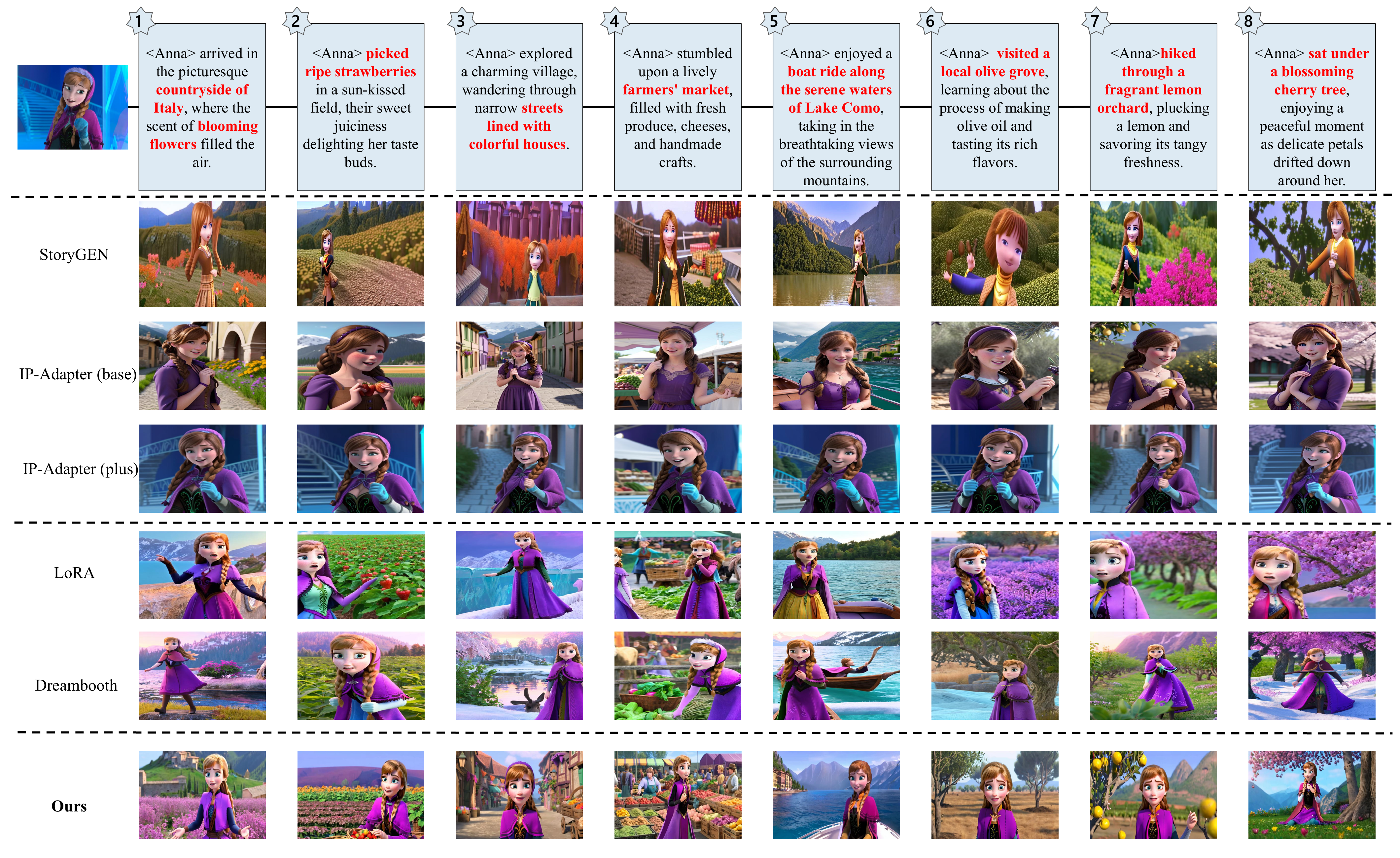} 
}
\subfigure[]{
\includegraphics[width=1.\linewidth]{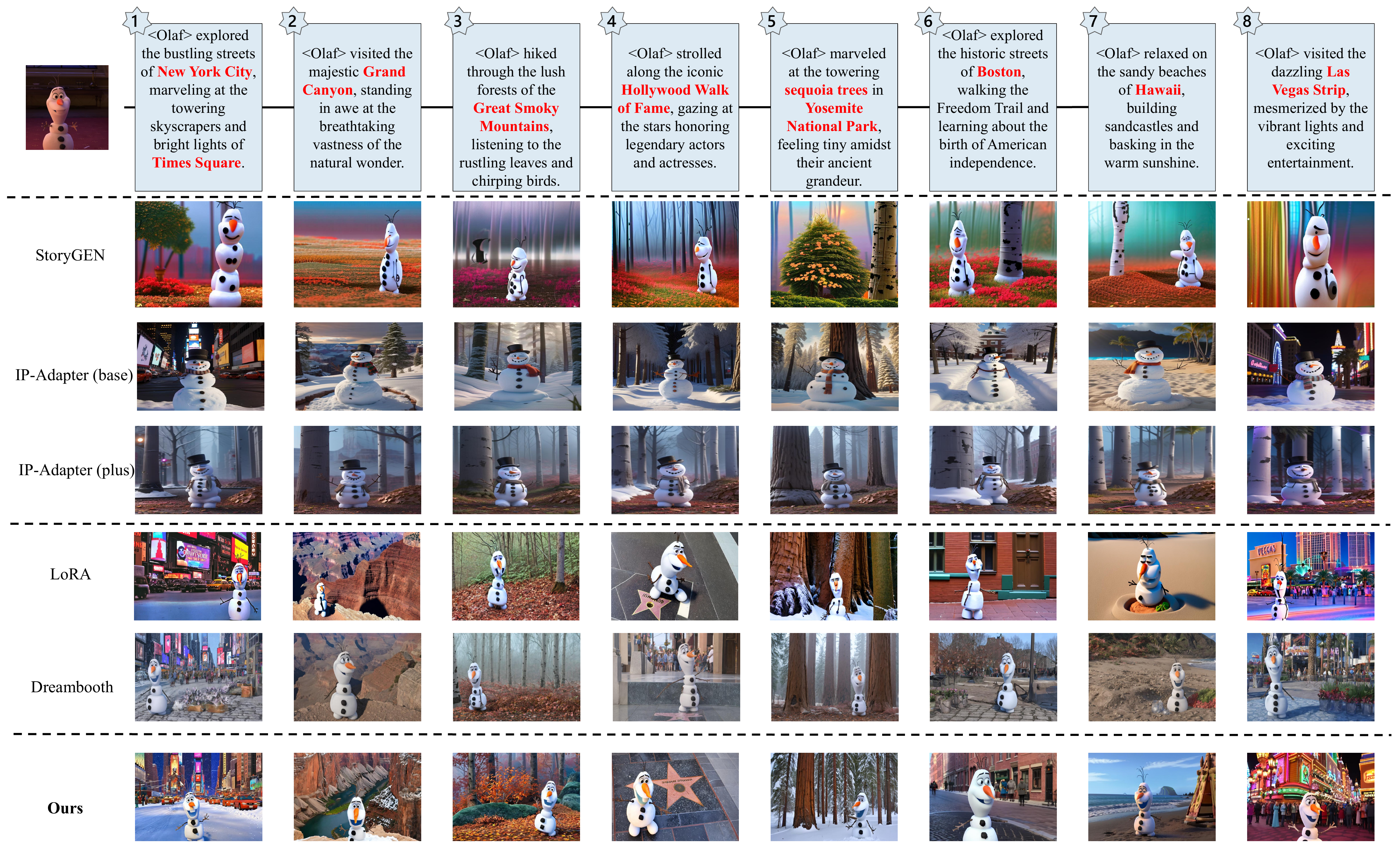} 
}
\vspace{-3mm}
\caption{Examples for single-character story visualization on Frozen Dataset.}
\label{fig:single-frozen}
\end{figure*}

\subsubsubsection{Single-Character Story Visualization}
In Fig. \ref{fig:single-pororo} and Fig. \ref{fig:single-frozen}, we present additional comparisons for single-character visual storytelling. The results depicted in these figures highlight the superior consistency in style and character exhibited by our proposed StoryWeaver, along with enhanced alignment between text and image.

\begin{figure*}[h]
\centering
\vspace{-5mm}
\subfigure[]{
\includegraphics[width=1.\linewidth]{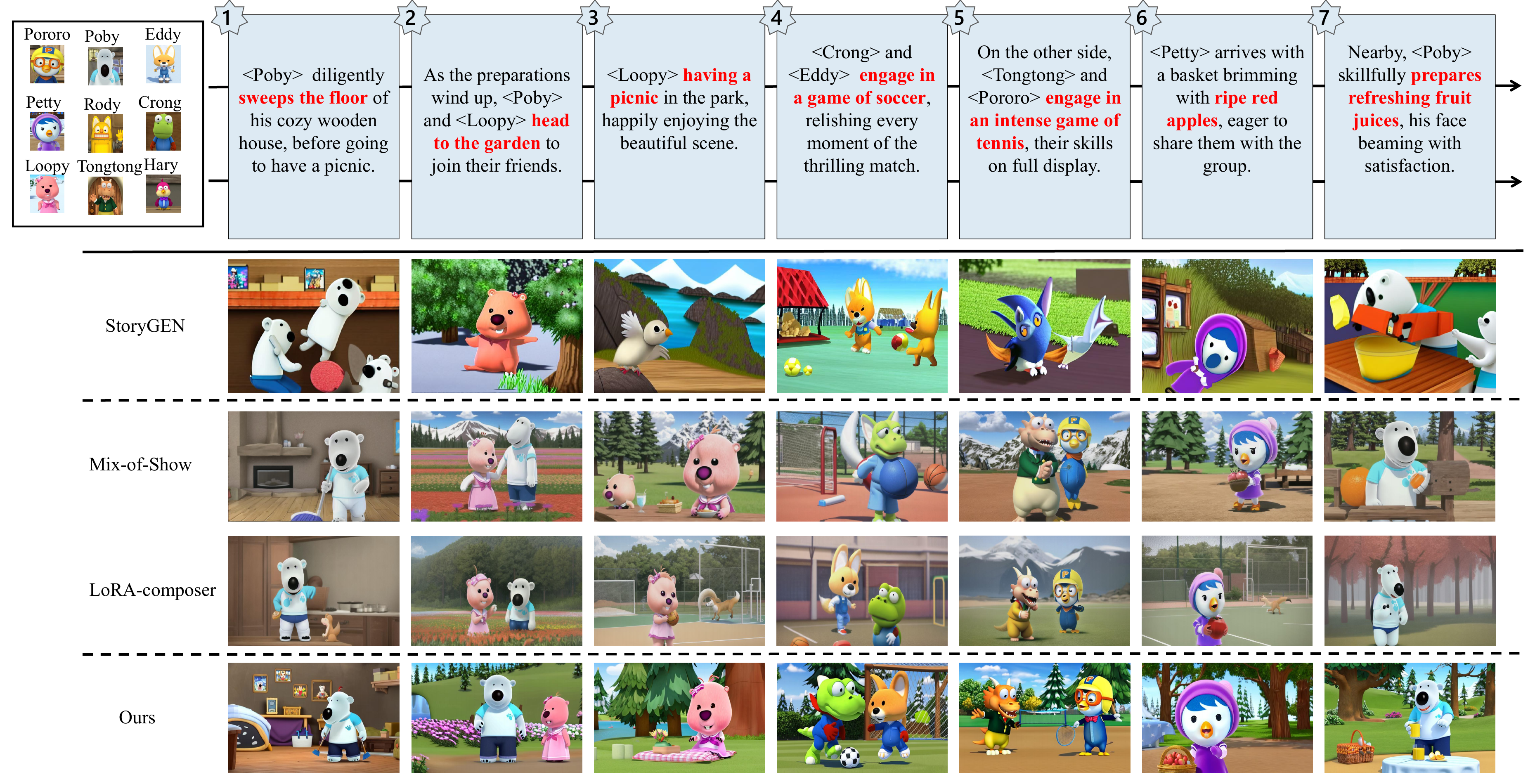} 
}
\subfigure[]{
\includegraphics[width=1.\linewidth]{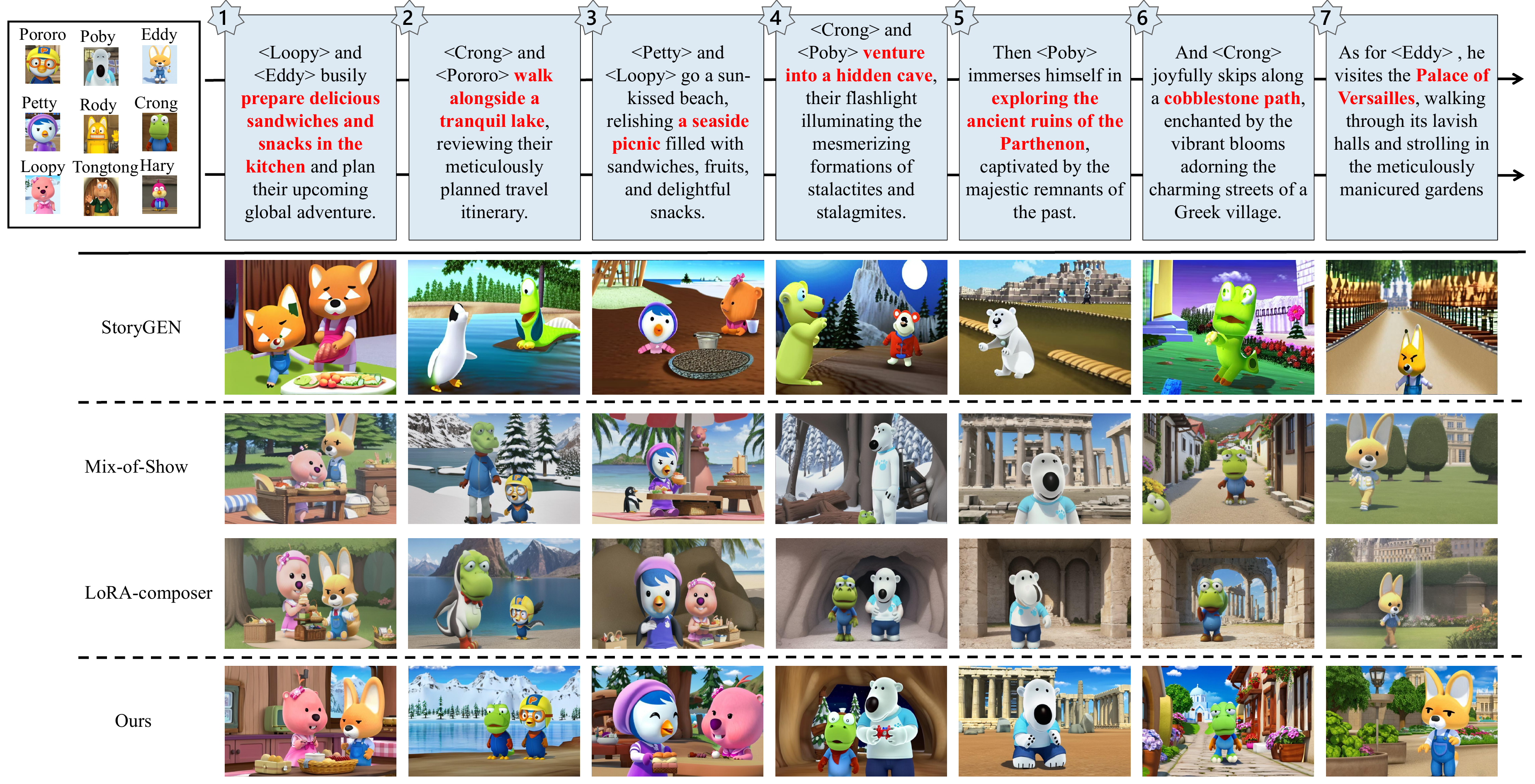} 
}
\vspace{-3mm}
\caption{Examples for multi-character story visualization on Pororo Dataset.}
\label{fig:multi-pororo}
\end{figure*}

\begin{figure*}[h]
\centering
\vspace{-5mm}
\subfigure[]{
\includegraphics[width=1.\linewidth]{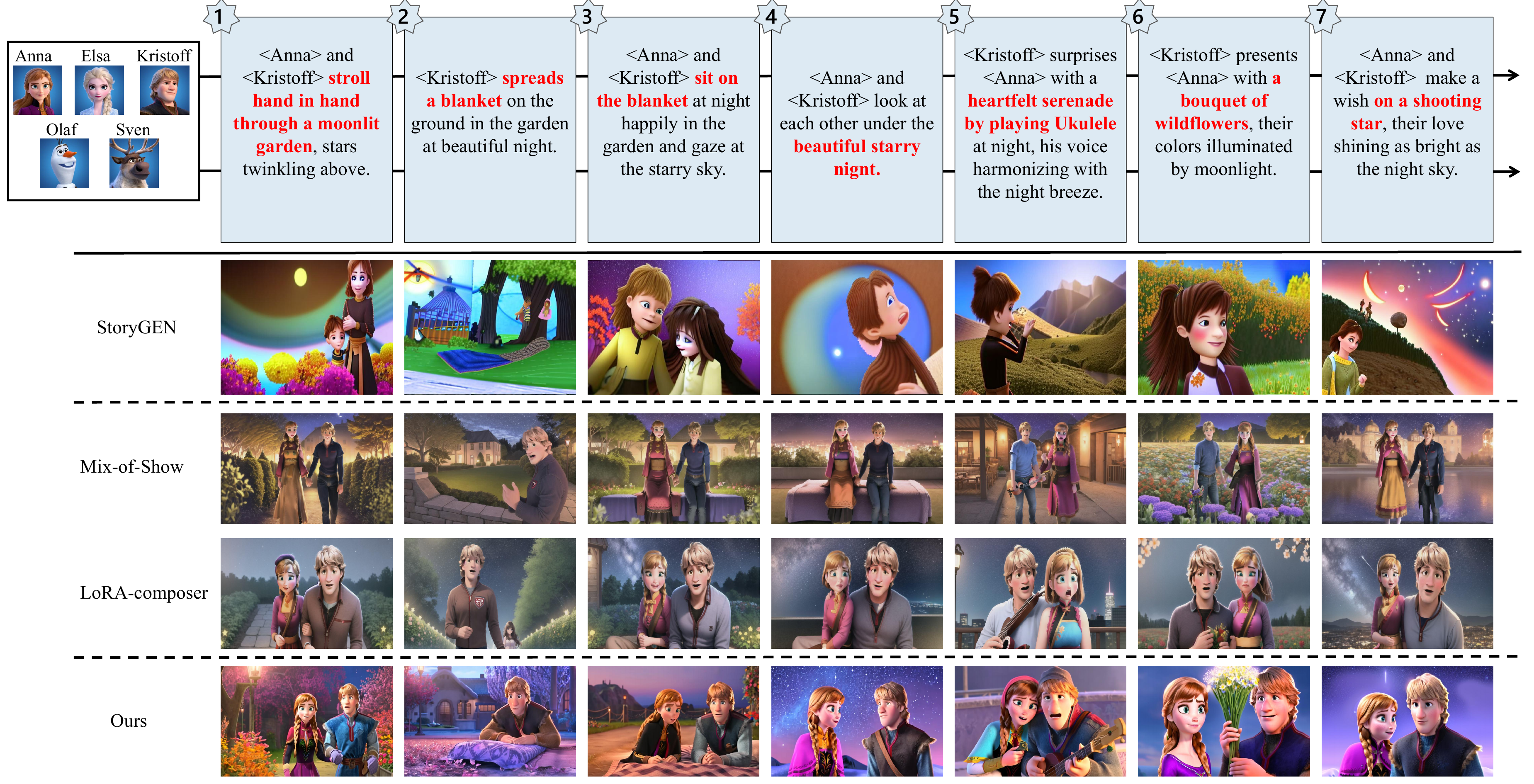} 
}
\subfigure[]{
\includegraphics[width=1.\linewidth]{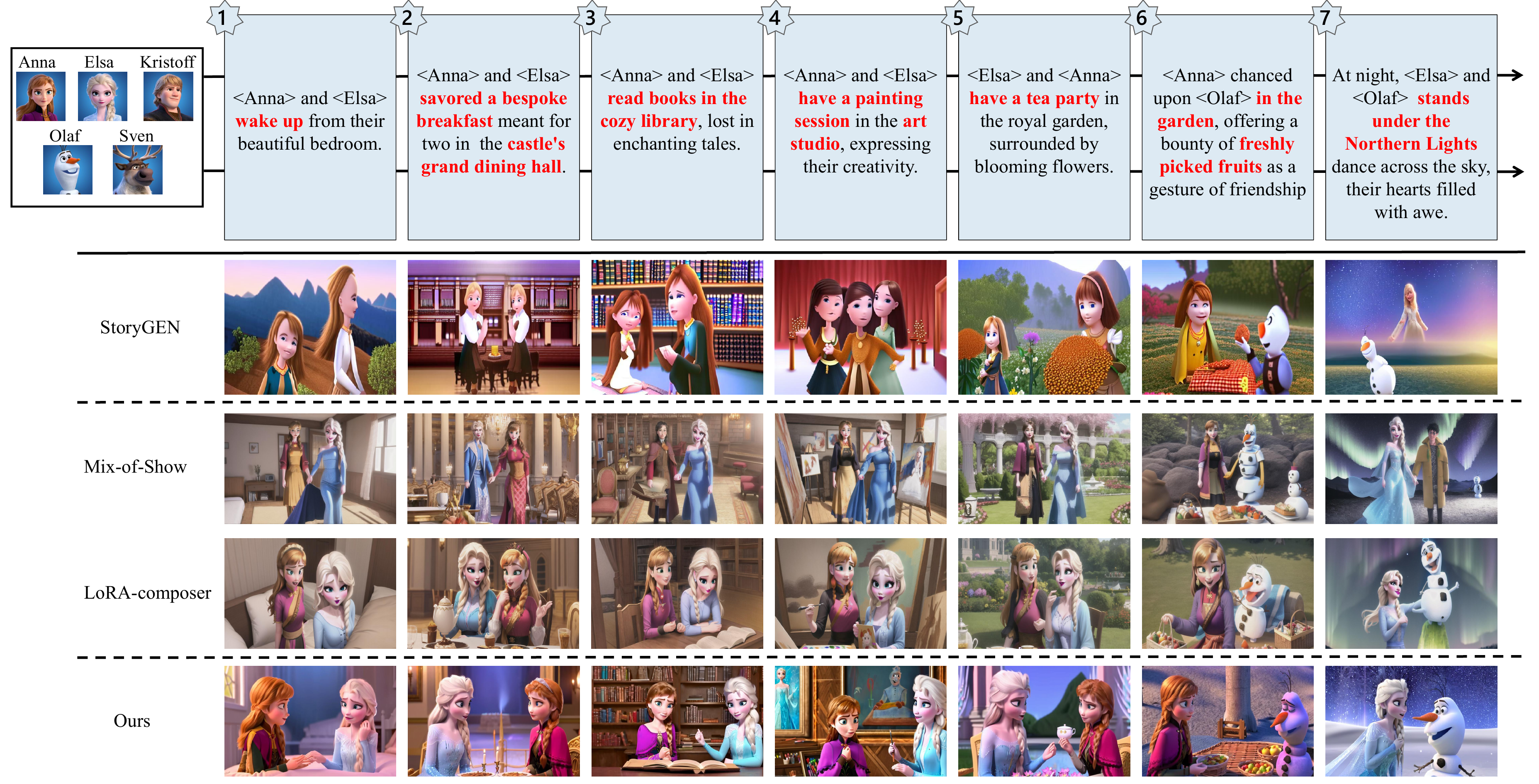} 
}
\vspace{-3mm}
\caption{Examples for multi-character story visualization on Frozen Dataset.}
\label{fig:multi-frozen}
\end{figure*}

\subsubsubsection{Multi-Character Story Visualization}
Further visualizations for multi-character visual storytelling are shown in Fig.\ref{fig:multi-pororo} and Fig.\ref{fig:multi-frozen}. In addition to the improved text-image alignment, StoryWeaver excels in generating natural interactions among characters while effectively preserving the identity of each character throughout the narrative frames.

\bigskip
